\documentclass{article} % For LaTeX2e
\usepackage{iclr2024_conference,times}

\usepackage{amsmath,amsfonts,bm}

\def\eqref#1{equation~\ref{#1}}
\def\1{\bm{1}}

\DeclareMathAlphabet{\mathsfit}{\encodingdefault}{\sfdefault}{m}{sl}
\SetMathAlphabet{\mathsfit}{bold}{\encodingdefault}{\sfdefault}{bx}{n}

\usepackage{hyperref}
\usepackage{url}

\usepackage[utf8]{inputenc} % allow utf-8 input
\usepackage[T1]{fontenc}    % use 8-bit T1 fonts
\usepackage{hyperref}       % hyperlinks
\usepackage{url}            % simple URL typesetting
\usepackage{booktabs}       % professional-quality tables
\usepackage{amsfonts}       % blackboard math symbols
\usepackage{nicefrac}       % compact symbols for 1/2, etc.
\usepackage{microtype}      % microtypography
\usepackage{xcolor}         % colors

\usepackage{caption, subcaption}
\usepackage{wrapfig}

\usepackage{amssymb, amsthm, amsmath, mathtools}
\renewcommand{\eqref}[1]{Eq.~(\ref{#1})}

\theoremstyle{plain}
\newtheorem{theorem}{Theorem}[section]

\theoremstyle{definition}
\newtheorem{definition}[theorem]{Definition}

\theoremstyle{remark}

\usepackage{mathtools}
\usepackage{bbm}
\usepackage{verbatim}
\usepackage{array,colortbl,multirow,multicol,booktabs}
\usepackage{textcomp}

\usepackage{algorithm}
\usepackage{algorithmic}

\DeclareMathOperator{\age}{age}
\DeclareMathOperator{\race}{race}

\DeclareMathOperator{\white}{white}

\usepackage{pifont}
\newcommand{\cmark}{\ding{51}}%
\newcommand{\xmark}{\ding{55}}%

\usepackage{url}

\definecolor{Gray}{gray}{0.9}

\title{Self-supervised debiasing using \\ low rank regularization}

\author{Geon Yeong Park$^{1}$, Chanyong Jung$^{1}$, Sangmin Lee$^{2}$, Jong Chul Ye$^{3,1*}$, Sang Wan Lee$^{1}\thanks{Co-corresponding authors.}$ \\
Bio and Brain Engineering$^{1}$, Mathematical Sciences$^{2}$, Kim Jaechul Graduate School of AI$^{3}$ \\ 
KAIST \\
\texttt{\{pky3436,jcy132,leeleesang,jong.ye,sangwan\}@kaist.ac.kr}
}

\begin{document}

\maketitle

\begin{abstract}
  Spurious correlations can cause strong biases in deep neural networks, impairing generalization ability. While most existing debiasing methods require full supervision on either spurious attributes or target labels, training a debiased model from a limited amount of both annotations is still an open question. To address this issue, we investigate an interesting phenomenon using the spectral analysis of latent representations: spuriously correlated attributes make neural networks inductively biased towards encoding lower effective rank representations. We also show that a rank regularization can amplify this bias in a way that encourages highly correlated features. Leveraging these findings, we propose a self-supervised debiasing framework potentially compatible with unlabeled samples. 
  Specifically, we first pretrain a biased encoder in a self-supervised manner with the rank regularization, serving as a semantic bottleneck to enforce the encoder to learn the spuriously correlated attributes. This biased encoder is then used to discover and upweight bias-conflicting samples in a downstream task, serving as a boosting to effectively debias the main model. Remarkably, the proposed debiasing framework significantly improves the generalization performance of self-supervised learning baselines and, in some cases, even outperforms state-of-the-art supervised debiasing approaches.
\end{abstract}

\section{Introduction}
\vspace{-0.2cm}
While modern deep learning solves several challenging tasks successfully, a series of recent works \citep{geirhos2018imagenet, gururangan2018annotation, feldman2015certifying} have reported that the high accuracy of deep networks on in-distribution samples does not always guarantee low test error on out-of-distribution (OOD) samples, especially in the context of spurious correlations. Existing studies \citep{arjovsky2019invariant, nagarajan2020understanding, tsipras2018robustness} suggest that the deep networks can be potentially biased to the spuriously correlated attributes, or dataset bias, which are misleading statistical heuristics that are closely correlated but not causally related to the target label. %In this regard, several recent works explain this phenomenon through the lens of simplicity bias \citep{rahaman2019spectral, neyshabur2014search, shah2020pitfalls} of gradient descent-based deep networks optimization; deep networks prefer to rely on spurious features which are more ``simpler" to learn, e.g., more linear.

These catastrophic pitfalls of dataset bias have facilitated the development of debiasing methods, which can be roughly categorized into approaches: (\textbf{1}) leveraging annotations of spurious attributes, i.e., bias label  \citep{kim2019learning, sagawa2019distributionally, wang2020towards, tartaglione2021end};  (\textbf{2}) presuming specific type of bias, e.g., color and texture \citep{bahng2020learning, wang2019learning, ge2021robust}; or (\textbf{3}) without using explicit kinds of supervisions on dataset bias \citep{liu2021just, nam2020learning, lee2021learning, levy2020large, zhang2022correct}.

While substantial advances have been made in this regard, these approaches still fail to address the problem: how to train a debiased classifier by fully exploiting unlabeled samples lacking \textit{both} bias and target label.
More specifically, while the large-scale unlabeled dataset can be potentially biased towards spuriously correlated sensitive attributes, e.g., ethnicity, gender, or age \citep{abid2021large, agarwal2021evaluating}, current existing debiasing frameworks are not designed to deal with this real-world unsupervised settings. Here we also confirm that most supervised debiasing frameworks suffer from performance degradation in the low-labeled data setting.
Moreover, recent works have suggested that self-supervised learning might not be sufficient to deal with OOD generalization \citep{geirhos2020surprising, chen2021intriguing, robinson2021can} when dataset bias remains after data augmentation.

To tackle this issue, we first make a series of empirical observations that allow us to examine the fundamental difference between biased and unbiased representations. 
Interestingly, we found that spurious correlations suppress the effective rank \citep{roy2007effective} of latent representations, which severely deteriorates the semantic diversity of representations and leads to the degradation of feature discriminability. 
Another notable aspect of our findings is that the intentional increase of feature redundancy amplifies “prejudice” in neural networks. To be specific, as we enforce the correlation among latent features to regularize the effective rank of representations (i.e., rank regularization), the accuracy on bias-conflicting samples quickly declines while the model still performs reasonably well on the bias-aligned \footnote{The \textit{bias-aligned} samples refer to data with a strong correlation between (potentially latent) spurious features and target labels. The \textit{bias-conflicting} samples refer to the opposite cases where spurious correlations do not exist.} samples.

Based on these observations, we propose a novel debiasing framework that can utilize both labeled \textit{and} unlabeled biased samples with rank regularization. The proposed method is fully compatible with both supervised and self-supervised scenarios, where such compatibility arises from the rank regularization that does \textit{not} rely on any labels. Specifically, for a supervised (self-supervised) setting, we train {1}) a biased classifier (encoder) with rank regularization, which serves as a semantic bottleneck limiting the semantic diversity of feature components, and {2}) the main classifier (encoder) with standard (self-)supervised learning approaches. The biased model affords us the leverage to uncover spurious correlations and identify bias-conflicting samples in a downstream task.

Our work is the first to unveil the bias-rank relationships and introduce an effective debiasing strategy to exploit potentially unlabeled data samples. We demonstrate the effectiveness of the proposed debiasing framework with various challenging real-world biased datasets, including MultiCMNIST \citep{li2022discover}, biased Chest X-ray databases, UTKFace, CelebA, etc., in both a supervised and self-supervised scenario. These experiments show that our method significantly outperforms other self-supervised baselines, and even state-of-the-art supervised debiasing methods in some cases. 
%In a supervised scenario, we show that the rank-regularized model can effectively discover the bias-conflicting samples in challenging datasets such as biased Chest X-ray databases or datasets with multiple bias attributes \citep{li2022discover}. Based on these observations, we found that the proposed framework significantly improves the OOD generalization in the linear evaluation protocol \citep{oord2018representation}, even without making any modifications on the pretrained encoder, or outperforms state-of-the-art supervised debiasing methods in a semi-supervised learning scenario.

%Large-scale experiments demonstrate that our method significantly outperforms other self-supervised baselines, and even state-of-the-art supervised debiasing methods in some cases.

\begin{figure}[t]
    \centering
    \includegraphics[width=0.9\textwidth]{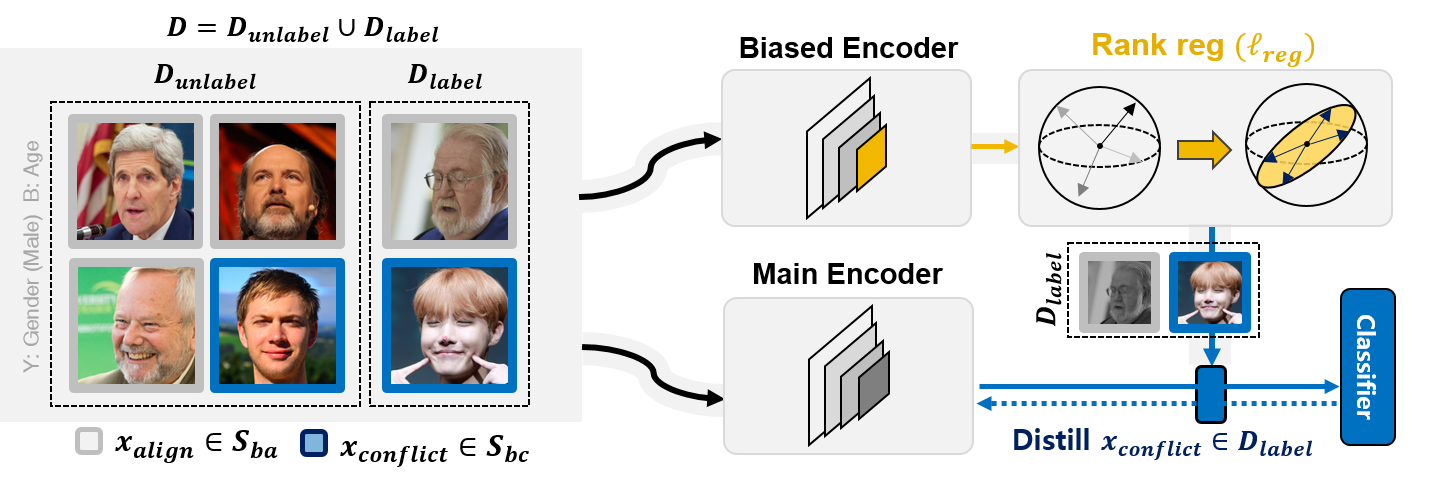}
    \caption{\textbf{Concept.} {Based on the bias-rank relationship (Section \ref{sec: low-rank}), we introduce a novel debiasing framework centered on rank regularization, which intentionally amplifies spurious correlation by enforcing feature components to be \textit{entangled} with both spurious and invariant attributes.}}
    \label{1 fig: concept}
\end{figure}

\vspace{-0.3cm}
\section{Low-rank bias of biased representations}
\label{sec: low-rank}
\vspace{-0.2cm}
\subsection{Preliminaries}
\label{sec: preliminaries}
\vspace{-0.2cm}

%\textbf{Notations.} 
Throughout the paper, we denote \(x \in \mathbb{R}^m\) and \(y \in \mathcal{Y}\) as \(m\)-dimensional input sample and its corresponding predicting label, respectively. Then we denote \(X=\{x_k\}_{k=1}^n\) as a batch of \(n\) samples from a dataset which is fed to an encoder \(f_\theta: \mathbb{R}^{m} \rightarrow \mathbb{R}^{d}\), parameterized by \(\theta\). Then we construct a matrix \(Z \in \mathbb{R}^{n \times d}\) where each \(i\)th row is the output representations of the encoder \(f_\theta(x_i)^T\) for \(x_i \in X\). For every analysis in this section, we use \(Z\) as our latent representations, where the neural backbone of the encoder may vary as simple convolutional networks, ResNet-18, or ViT \citep{dosovitskiy2020image} (Experimental details provided in Appendix \ref{sec: more observations} and \ref{sec: experimental setup}).

To evaluate the semantic diversity of given representation matrix, we introduce \textit{effective rank} \citep{roy2007effective} which is a widely used metric to measure the effective dimensionality of matrix and analyze the spectral properties of features in neural networks \citep{arora2019implicit, razin2020implicit, huh2021low, baratin2021implicit}:

\begin{definition}
\label{2 def: effective rank}
Given the matrix \(X \in \mathbb{R}^{m \times n}\) and its singular values \(\{\sigma_i\}_{i=1}^{\min{(m, n)}}\), the effective rank \(\rho\) of \(X\) is defined as the shannon entropy of normalized singular values: \vspace{-0.1cm}
\begin{align} \label{2 eq: effective rank} 
\rho(X) = - \sum_{i=1}^{\min{(m, n)}} \bar{\sigma_i} \log \bar{\sigma_i}, 
\end{align}
where \(\bar{\sigma_i} = \sigma_i / \sum_k \sigma_k\) is \(i\)-th normalized singular value. Without loss of generality, we omit the exponentiation of \(\rho(X)\) as done in \citep{roy2007effective}.
\end{definition}

% While the rank of given matrix represents the number of dimensions retained by the transformation, i.e., the dimension of its range, it does not provide detailed information about the induced geometrical shaping \citep{roy2007effective}. However, effective rank quantifies such geometrical transformations in terms of spectral entropy. Thus it can be regarded as a metric of effective dimensionality of matrix. 

\vspace{-0.2cm}
Effective rank is also referred to as spectral entropy where its value is maximized when the singular values are all equal and minimized when a top singular value dominates relative to all others. Recent works \citep{chen2019transferability, chen2019catastrophic} have revealed that the discriminability of representations resides on wide range of eigenvectors since the rich discriminative information for the classification task cannot be transmitted by only few eigenvectors with top singular values. Thus from a spectral analysis perspective, effective rank quantifies how diverse the semantic information encoded by each eigenfeature is, which is closely related to the feature discriminability across target label categories. In the rest of paper, we interchangeably use effective rank and rank by following prior works.

\begin{figure}[!t]
\centering
\begin{subfigure}[c]{0.21\textwidth}
\includegraphics[width=\textwidth]{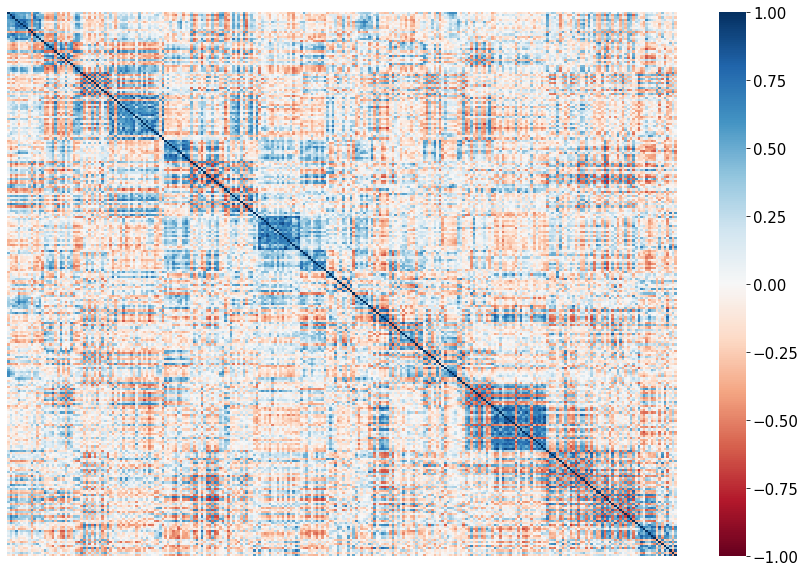} 
\caption[c]{Unbiased corr.}
\label{2 fig: unbias corr}
\end{subfigure}
\hfill
\begin{subfigure}[c]{0.2\textwidth}
\includegraphics[width=\textwidth]{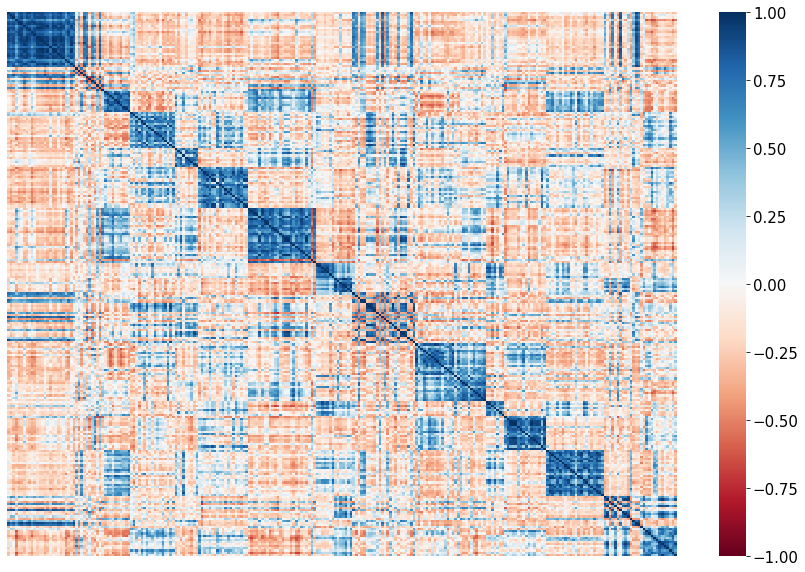}
\caption[c]{Biased corr.}
\label{2 fig: bias corr}
\end{subfigure}
\hfill
\begin{subfigure}[c]{0.21\textwidth}
\includegraphics[width=\textwidth]{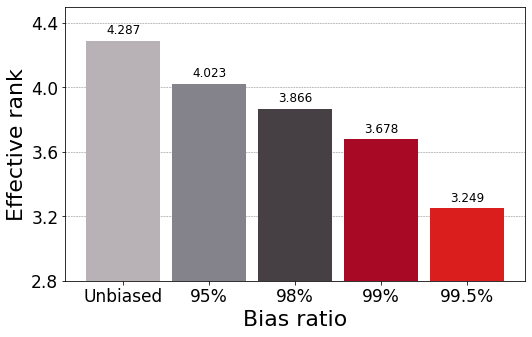}
\caption[c]{Effective rank}
\label{2 fig: cmnist rank}
\end{subfigure}
\hfill
\begin{subfigure}[c]{0.25\textwidth}
\includegraphics[width=\textwidth]{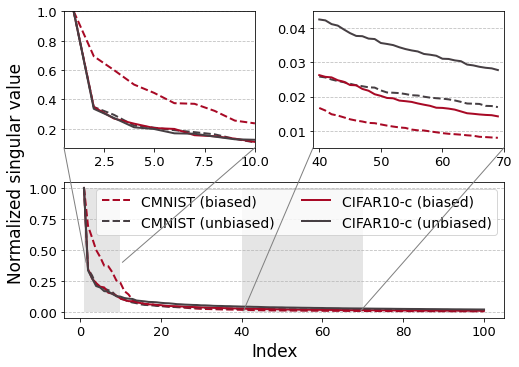}
\caption[c]{SVD analysis}
\label{2 fig: lambda}
\end{subfigure}
\caption{Empirical analysis on rank reduction phenomenon. For every analysis, we used the output $Z$ of the encoder (Sec. \ref{sec: preliminaries}). (\textbf{a}, \textbf{b}): Hierarchically clustered auto-correlation matrix of unbiased and biased representations (Bias ratio=99\(\%\)). (\textbf{c}): Effective rank with color bias. `Unbiased' represents the case training model with perfectly unbiased dataset, i.e. random color for each sample. (\textbf{d}): SVD analysis with max-normalized singular values. Top 100 values are shown in the figure (total: 256).
\vspace{-0.5cm}
} 
\end{figure}
\begin{comment}
\begin{figure}[t]
\centering
\includegraphics[width=0.4\textwidth]{figures/lambda_normalized_cmnist_cifar.png}
\caption[c]{SVD analysis with max-normalized singular values. Top 100 values are shown in the figure (Total: 256).}
\label{2 fig: lambda}
\end{figure}
\end{comment}

% \newpage \ 
\vspace{-0.15cm}
\subsection{Spectral analysis of the bias-rank relationships}
\vspace{-0.1cm}
We now present experiments showing that the deep networks may tend to encode lower-rank representations in the presence of stronger spurious correlations. 
%\textbf{Rank reduction phenomenon.} 
To arbitrarily control the degree of spurious correlations, we introduce synthetic biased datasets, Color-MNIST (CMNIST) and Corrupted CIFAR-10 (CIFAR-10C, \citep{hendrycks2019benchmarking}), with color and corruption bias types, respectively. We define the degree of spurious correlations as the ratio of bias-aligned samples included in the training set, or bias ratio, where most of the samples are bias-aligned in the context of strong spurious correlations. 

Figure \ref{2 fig: cmnist rank} shows that the rank of latent representations from a penultimate layer of the simple convolutional classifier decreases as the bias ratio increases in CMNIST. We provide similar rank reduction results of CIFAR-10C with ResNet-18 and ViT in the Appendix \ref{sec: more observations}. We further compare the correlation matrix of biased and unbiased latent representations in the penultimate layer of biased and unbiased classifiers, respectively. In Figure \ref{2 fig: unbias corr} and \ref{2 fig: bias corr}, we observe that the block structure in the correlation matrix is more evident in the biased representations after the hierarchical clustering, indicating that the features become highly correlated which may limit the semantic diversity of networks. To investigate the rank reduction phenomenon in-depth, we compare the normalized singular values of biased and unbiased representations. We conduct singular value decomposition (SVD) on the feature matrices of both biased and unbiased classifiers and plot the singular values normalized by the spectral norm of the corresponding matrix. Figure \ref{2 fig: lambda} shows that the top few normalized singular values of biased representations are similar to or even greater than those of unbiased representations. However, the remaining majority of singular values decay significantly faster in biased representations, greatly weakening the informative signals of eigenvectors with smaller singular values and deteriorating feature discriminability \citep{chen2019transferability, chen2019catastrophic}.

\vspace{-0.3cm}
\subsection{Rank regularization} \label{sec: rank regularization}
\vspace{-0.2cm}

Motivated from the aforementioned rank reduction phenomenon, we ask an opposite-directional question: ``Can we intentionally amplify the prejudice of deep networks by \textit{maximizing} the redundancy between the components of latent representations?". If the feature components are extremely correlated, the corresponding representations may exhibit most of its spectral energy along the direction of one singular vector. For this case, effective rank may converge to 0. In other words, our goal is to design a \textit{semantic bottleneck} of representations that restricts the semantic diversity of feature vectors. To implement the bottleneck in practice, motivated from Figure \ref{2 fig: bias corr}, we compute the auto-correlation matrix of the output of encoder. 

Let \(\bar{Z}\) denote the mean-centered representations \(Z\) along the batch dimension. The normalized auto-correlation matrix \(C \in \mathbb{R}^{d \times d}\) of \(\bar{Z}\) is defined as follow:
\vspace{-0.1cm}
\begin{align} \label{2.2 eq: auto correlation}
    C_{i, j} = \frac{\sum_{b=1}^{n} \bar{Z}_{b, i} \bar{Z}_{b, j}}{\sqrt{\sum_{b=1}^n \bar{Z}_{b, i}^2} {\sqrt{\sum_{b=1}^n \bar{Z}_{b, j}^2}}} \quad 1 \leq \forall i, j \leq d,
\end{align}
% \vspace{-0.2cm}
where \(b\) is an index of sample and \(i, j\) are index of each vector dimension. Then we define our regularization term as the negative of a sum of squared off-diagonal terms in \(C\):

\vspace{-0.2cm}
\begin{align}
\label{2.2 eq: rank reg}
\ell_{reg}(X; \theta) = - \sum_{i} \sum_{j\neq i} C_{i,j}^2,
\end{align}
where we refer to it as the {\em rank loss}. Note that the target labels on \(X\) is \textit{not} used at all. 
\vspace{-0.2cm}
%While the SVD operation for (\ref{2 eq: effective rank}) is computationally expensive, the proposed regularization term is relatively efficient and easy to implement. 

\begin{figure}[t]
\centering
\begin{subfigure}[c]{0.3\textwidth}
\includegraphics[width=\textwidth]{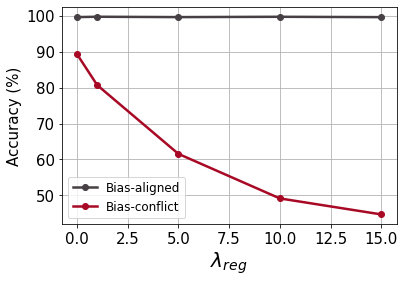} 
\subcaption[c]{CMNIST}
\label{2 fig: cmnist group}
\end{subfigure}
\hfill
\begin{subfigure}[c]{0.3\textwidth}
\includegraphics[width=\textwidth]{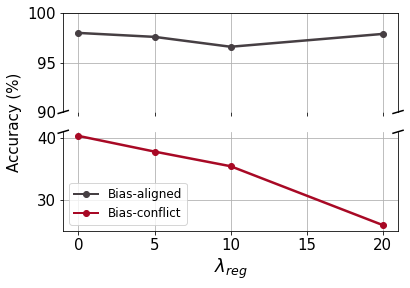}
\subcaption[c]{CIFAR-10C}
\label{2 fig: cifar group}
\end{subfigure}
\hfill
\begin{subfigure}[c]{0.3\textwidth}
\includegraphics[width=\textwidth]{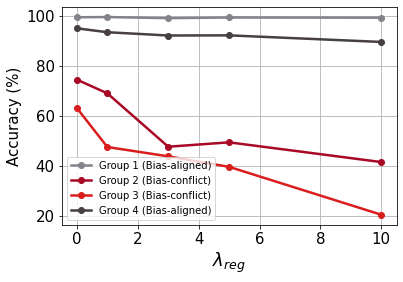}
\subcaption[c]{Waterbirds}
\label{2 fig: waterbirds group}
\end{subfigure}
\hfill
\caption{(\textbf{a}, \textbf{b}): Bias-conflict and Bias-aligned accuracy on CMNIST and CIFAR-10C (Bias ratio=95$\%$). (\textbf{c}): Group accuracy on Waterbirds. Detailed simulation settings are in the Appendix \ref{sec: experimental setup}.
\vspace{-0.25cm}
} 
\label{2 fig: rank reg}
\end{figure}

\begin{table}[tbp]
\centering
    \begin{subtable}[h]{0.48\textwidth}
        \centering
        \small
        \begin{tabular}{c c c c c}
            \toprule
            \multicolumn{1}{c}{\multirow{2}{*}{}} & \multicolumn{2}{c}{CMNIST} & \multicolumn{2}{c}{CIFAR-10C} \\
            \cmidrule{2-5}
            \multicolumn{1}{c}{} & P ($\%$) & R ($\%$) & P ($\%$) & R ($\%$) \\
            \midrule
            {ERM} & {85.59} & {19.76} & {52.03} & {0.06} \\
            \midrule
            {+ Rank reg} & {\textbf{98.83}} & {\textbf{95.91}} & {\textbf{71.39}} & {\textbf{51.43}} \\
            \bottomrule
       \end{tabular}
       \caption{CMNIST, CIFAR-10C}
       \label{2 table: precision}
    \end{subtable}
    \begin{subtable}[h]{0.48\textwidth}
        \centering
        \small
        \begin{tabular}{c c c c}
            \toprule
            {Metrics} & {ERM} & {JTT} & {Rank reg} \\
            \midrule
            {Precision ($\%$)} & {37.84} & {48.95} & {\textbf{54.77}} \\
            \midrule
            {Recall ($\%$)} & {11.67} & {48.75} & {\textbf{55.01}} \\
            \bottomrule
       \end{tabular}
       \caption{Waterbirds} 
       \label{2 table: precision_jtt}
    \end{subtable}
    \caption{Precision (P) and Recall (R) of bias-conflicting samples. (\textbf{a}): Bias-conflicting samples are identified in the error set of ERM model trained with and without rank regularization (Bias ratio=95$\%$ for both datasets). (\textbf{b}): Bias-conflicting samples are similarly identified by ERM, JTT, and the proposed biased model in Waterbirds dataset.}
\end{table}

\textbf{Analysis of rank-regularized networks.} To investigate the impacts of rank regularization in deep neural networks, we construct the classification model by combining the linear classifier \(f_W: \mathbb{R}^d \rightarrow \mathbb{R}^c\) parameterized by \(W \in \mathcal{W}\) on top of the encoder \(f_\theta\), where \(c = |\mathcal{Y}|\) is the number of classes. Then we trained models by cross entropy loss \(\ell_{CE}\) combined with \(\lambda_{reg} \ell_{reg}\), where \(\lambda_{reg} > 0\) is a Lagrangian multiplier. We use CMNIST, CIFAR-10C, and Waterbirds dataset \citep{wah2011caltech}, and evaluate the trained models on an unbiased test set following \cite{nam2020learning, lee2021learning}. After training models with varying the hyperparameter \(\lambda_{reg}\), we compare bias-aligned and bias-conflict accuracy, which are the average accuracy on bias-aligned and bias-conflicting samples in the unbiased test set, respectively, for CMNIST and CIFAR-10C. Test accuracy on every individual data group is reported for Waterbirds. 
Figure \ref{2 fig: rank reg} shows that models suffer more from poor OOD generalization as trained with larger \(\lambda_{reg}\). The average accuracy on bias-conflicting groups is significantly degraded, while the accuracy on bias-aligned groups is maintained to some extent. It implies that rank regularization may force deep networks to focus on spurious attributes. 

\textbf{Minority mining performance.} Table \ref{2 table: precision} and \ref{2 table: precision_jtt} support that the biased models with strong regularization can effectively probe out the bias-conflicting samples in the training set. Specifically, we train a biased classifier with rank regularization and distill an error set \(E\) of misclassified training samples as bias-conflicting samples proxies. As reported in Table \ref{2 table: precision}, we observe that our biased classifier is relatively robust to the unintended memorization of bias-conflicting samples \citep{sagawa2020investigation} in contrast to the standard models trained by Empirical Risk Minimization (ERM). Moreover, Table \ref{2 table: precision_jtt} shows that the proposed rank regularization improves the precision and recall of identified bias-conflicting samples compared to JTT \citep{liu2021just}. Detailed simulation settings are in the Appendix \ref{sec: experimental setup}.
% Note that we did not use any annotations .
\begin{figure*}[!t]
\centering
\begin{subfigure}[c]{0.279\textwidth}
\includegraphics[width=\textwidth]{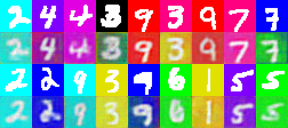} 
\subcaption[c]{Unbiased}
\label{2 fig: unbiased recon}
\end{subfigure}
\begin{subfigure}[c]{0.279\textwidth}
\includegraphics[width=\textwidth]{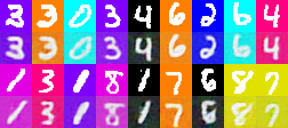} 
\subcaption[c]{Biased}
\label{2 fig: biased recon}
\end{subfigure}
\begin{subfigure}[c]{0.279\textwidth}
\includegraphics[width=\textwidth]{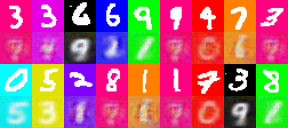} 
\subcaption[c]{Biased with rank reg.}
\label{2 fig: biased recon rank reg}
\end{subfigure}
\begin{subfigure}[c]{0.142\textwidth}
\includegraphics[width=\textwidth]{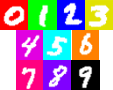} 
\subcaption[c]{Examples.}
\label{2 fig: cmnist_aligned}
\end{subfigure}
\caption{Randomly selected reconstructed images from representations with varying degrees of bias. First and third row correspond to the input bias-conflicting images. Second and fourth row correspond to the reconstructed images. Reconstructed from (\textbf{a}) unbiased representations, (\textbf{b}) biased representations, and (\textbf{c}) biased representations with rank regularization (bias ratio=95$\%$ in \textbf{b, c}). (\textbf{d}) Examples of bias-aligned CMNIST images.
\vspace{-0.25cm}
}
\label{2 fig: recon}
\end{figure*}

%\begin{wrapfigure}[6]{r}{0.3\textwidth}
%\centering
%\includegraphics[width=0.29\textwidth]{figures/cmnist_example2.png}
%\caption{Examples of bias-aligned CMNIST images.}\label{2 fig: cmnist_aligned}
%\end{wrapfigure}

\textbf{Reconstruction of biased representations.} To understand the relationship between rank regularization and spurious correlations more deeply, we visualize the pretrained representations with varying degrees of bias. We first trained deep networks on: (\textbf{a}) unbiased CMNIST (random background color), (\textbf{b}) biased CMNIST (bias ratio=95$\%$) without rank regularization and (\textbf{c}) with rank regularization ($\lambda_{reg}=50$). Then, we train the auxiliary decoder, which reconstructs the bias-conflicting images from the freezed latent representations of each pretrained network.
Results show that rank regularization may cause the representation to lose information on complex invariant features, resulting in a loss of feature discriminability and informative signals. While both digit and color are well reconstructed with biased representations (\textbf{b}), the decoder fails to reconstruct bias-conflicting images from the (\textbf{c}) biased representations pretrained with rank regularization. The foreground digit is blurred, and its class changes following the color-digit assignment in Figure \ref{2 fig: cmnist_aligned}.

These observations afford us some key insights into rank regularization: First, the rank-regularized representation may lose its information on complex invariant features (i.e., shape and style of the foreground digit), specifically undermining the feature discriminability and informative signals. Second, the limited semantic diversity makes it harder to identify the true underlying independent generative factors for multidimensional data; instead, it may encode feature components \textit{entangled} with both spurious and invariant attributes as the digit class of the reconstructed image is erroneously determined by the background color in \ref{2 fig: biased recon rank reg}.

\textbf{Multiple bias attributes.} To further investigate the generalizability of rank regularization, we evaluate the biased representations with Multi-Color MNIST (MultiCMNIST) dataset \citep{li2022discover}, which is similar to the CMNIST but have two bias attributes: left and right background colors. We set bias ratio=99$\%$ for the left color and bias ratio=95$\%$ for the right color, i.e., the left color is a more salient bias than the right color (Dataset details are provided in Appendix \ref{sec: experimental setup}).

\begin{table}[!hbt]
\caption{(\textbf{a}) Test accuracy ($\%$) on MultiCMNIST. Lower is better for this results. BC for bias-conflicting, and BA for bias-aligned. Bias ratio=99($\%$) for left color, and 95($\%$) for right color. $\lambda_{reg}=50$ is used for rank regularization. (\textbf{b}) Debiasing results. Higher is better for this results. Baseline results are from \cite{li2022discover}. $\lambda_{up}=50$ is used for upweighting in the proposed framework ($\lambda_{up}$: a manual rescaling weight given to each identified bias-conflicting samples in cross entropy loss). Pseudo-code and experimental details are provided in Appendix \ref{sec: pseudocode} and \ref{sec: experimental setup}, respectively.}
\label{2. table: MultiCMNIST}
\centering
\resizebox{0.99\textwidth}{!}{
\begin{tabular}{c c c c c c c c c c}
\toprule
\multicolumn{1}{c}{\multirow{2}{*}{Idx}} & \multicolumn{1}{c}{\multirow{2}{*}{Left color}} & \multicolumn{1}{c}{\multirow{2}{*}{Right color}} & \multicolumn{3}{c}{(\textbf{a}) Biased accuracy ($\%$)} & \multicolumn{3}{c}{(\textbf{b}) Debiased accuracy ($\%$)} & \\
\cmidrule{4-9}
\multicolumn{1}{c}{} & \multicolumn{1}{c}{} & \multicolumn{1}{c}{} & ERM & LfF \citep{nam2020learning} & \textbf{Rank reg.} & LfF & DebiAN \citep{li2022discover} & \textbf{Ours} \\
\midrule 
{(1)} & {BA} & {BC} & {100.0} & {100.0} & {100.0} & {99.6} & {100.0} & {100.0} \\ 
\midrule 
{(2)} & {BA} & {BC} & {96.6} & {98.8} & {41.6} & {4.7} & {95.6} & {97.0} \\
\midrule
{(3)} & {BC} & {BA} & {29.3} & {3.2} & {8.7} & {98.6} & {76.5} & {79.1} \\
\midrule
{(4)} & {BC} & {BC} & {7.6} & {1.3} & {6.1} & {5.1} & {16.0} & {18.3} \\
\midrule
\multicolumn{3}{c}{(1) $\sim$ (4) average acc.} & {58.38} & {50.83} & {\textbf{39.1}} & {52.0} & {72.0} & {\textbf{73.6}} \\
\bottomrule
\end{tabular}
}
\end{table}

Table \ref{2. table: MultiCMNIST} shows that the rank regularization successfully biases the model w.r.t both bias attributes, while LfF \citep{nam2020learning} completely fails to amplify the right color bias, i.e. less salient bias, as shown in the second row (Biased accuracy part). This leads to the abnormal debiasing results of LfF as shown in Table \ref{2. table: MultiCMNIST} where it records unbalanced accuracy for the left- and right-color-bias-conflicting samples. In contrast, the proposed framework shows superior performance by simply upweighting the misclassified bias-conflicting proxies, as done in \cite{liu2021just}.

Taken together, these results indicate that the rank regularization encourages the network to focus more on spurious correlations in a way that minimizes semantic diversity and \textit{entangles} invariant and spurious features \citep{park2023training}, which is a fundamentally different mechanism compared to the LfF \citep{nam2020learning} with its easy-to-learn assumption. More details on the upweighting strategy will be provided in Section \ref{sec: defund} and pseudo-code in Appendix \ref{sec: pseudocode}.
%In other words, the proposed low-rank regularization prevents features from encoding discriminative information independently. 

\vspace{-0.2cm}
\section{DeFund: Debiasing framework with unlabeled data}
\label{sec: defund}
\vspace{-0.2cm}
{Motivated by the observations in Section \ref{sec: low-rank}, we propose a self-supervised debiasing framework with unlabeled data, coined DeFund
(Debiasing Framework with Unlabeled Data). A notable distinction from previous studies \citep{bahng2020learning, zhang2022correct} lies in the proposed framework's ability to effectively harness unlabeled data for learning biased representations. This is achieved through the application of self-supervised learning and rank regularization techniques.}

%\textbf{Outline.} 
The proposed framework is composed of two stages: We first train the biased encoder, which can be potentially adopted to detect the bias-conflicting samples in a downstream task, along with the main encoder by self-supervised learning, both without any labels. After pretraining, we identify the bias-conflicting samples in the downstream task using linear evaluation protocol \citep{oord2018representation, chen2020simple}. This set of samples serves as a boosting to debias the main model.  

\textbf{Notation.} 
We denote \(f^{bias}_{\theta}: \mathcal{X} \rightarrow \mathbb{R}^d\) and \(f^{main}_{\phi}: \mathcal{X} \rightarrow \mathbb{R}^d\) as biased encoder and main encoder parameterzied by \(\theta \in \Theta\) and \(\phi \in \Theta\), respectively, where \(d\) is the dimensionality of latent representations. Then we can compute the rank loss in (\ref{2.2 eq: rank reg}) with introduced encoders and given batch \(\{x_k\}_{k=1}^N\) with size \(N\). Let \(f^{cls}_{W_b}: \mathbb{R}^{d} \rightarrow \mathbb{R}^C\) be a single-layer classifier parameterized by \(W_b \in \mathcal{W}\) which is placed on top of biased encoder \(f^{bias}_\theta\), where \(C = |\mathcal{Y}|\) is the number of classes. We similarly define the linear classifier \(f^{cls}_{W_m}\) for the main encoder. Then we refer to \(f^{bias}: \mathcal{X} \rightarrow \mathbb{R}^C\) as biased model, where \(f^{bias}(x) = f^{cls}_{W_b}\big( f^{bias}_{\theta}(x) \big), \forall x \in \mathcal{X}\). We similarly define the main model \(f^{main}\) as \(f^{main}(x) = f^{cls}_{W_m}\big( f^{main}_{\phi}(x) \big), \forall x \in \mathcal{X}\). While the projection networks \citep{chen2020simple} are employed as well, we omit the notations because they are not engaged in classification.

\textbf{Stage 1. Training a biased encoder.}  To train the biased encoder \(f^{bias}_\theta\), we revisit the proposed rank regularization term (\ref{2.2 eq: rank reg}) in context of instance discrimination task. Building upon the observations in Section \ref{sec: rank regularization}, we conjecture that rank regularization may amplify bias in self-supervised learning as well by entangling invariant and spurious features. Based on these intuitions, we apply rank regularization directly to the output of the base encoder, which encourages each feature component to be highly correlated. {From these applications, several noteworthy observations have emerged: (\textbf{a}) The representation becomes more biased as it is trained with stronger regularization (Appendix \ref{sec: more observations}). (\textbf{b}) While the overall performance may be upper-bounded due to the constraint on effective dimensionality \citep{jing2021understanding}, the bias-conflict accuracy is primarily sacrificed compared to the bias-aligned accuracy (Section \ref{sec: results})}.

\textbf{Stage 2. Debiasing downstream tasks.} {After training the biased encoder, our goal is to debias the main model, which was pretrained using standard self-supervised learning methods on the same dataset. Here, assume that we have an ideal pretrained main encoder of which each output component corresponds to the latent factor of data variation \citep{zimmermann2021contrastive}. While this ideal encoder should seamlessly adapt to downstream classification tasks, if most downstream task samples are bias-aligned, they may misguide the model to upweight spuriously correlated latent factors, leading to a biased solution despite well-generalized representations. We refer to this problem as the biased downstream application (Related analysis in Appendix \ref{sec: more observations})}.

The above contradiction elucidates the importance of bias-conflicting samples, which serve as counterexamples of spuriously correlated feature components, thereby preventing the alleged involvement of such components in prediction. Based on these intuitions, we introduce a novel debiasing protocol that probes and upweights bias-conflicting samples to find and fully exploit feature components independent of spurious correlations. We apply our framework on two scenarios: linear evaluation and semi-supervised learning. 

\textbf{Linear evaluation.} To validate our hypothesis on the biased downstream application, we conduct linear evaluation \citep{zhang2016colorful, oord2018representation} following the conventional protocol of self-supervised learning. Specifically, a linear classifier is trained on top of unsupervised pretrained representations by using target labels of training samples. After training a linear classifier \(f^{cls}_{W_b}\) with pretrained biased encoder \(f^{bias}_{\theta}\) given the whole training set \(D=\{(x_k, y_k)\}_{k=1}^{N}\) with size \(N\), an error set \(E\) of misclassified samples and corresponding labels is regarded as bias-conflicting pairs. Then we train a linear classifier \(f^{cls}_{W_m}\) on intentionally freezed representations of main encoder \(f^{main}_{\phi}\) by upweighting the identified samples in \(E\) with \(\lambda_{up} > 0\). The loss function for \textit{debiased} linear evaluation is defined as follows:
\begin{align*}
% \label{3 eq: upweighted loss}
% \begin{split}
\ell_{debias}(D; W_m) = \lambda_{up} \sum_{(x, y) \in E} \ell (x, y; W_m) + \sum_{(x, y) \in D \setminus E} \ell (x, y; W_m),
% \end{split}
\end{align*}
\vspace{-0.2cm}

where we use cross entropy loss for \(\ell: \mathcal{X} \times \mathcal{Y} \times \mathcal{W} \rightarrow \mathbb{R}^{+}\). Note that the target labels are only used in training linear classifiers after pretraining.

%\add{\textbf{Remark.} 
Note that the debiased linear evaluation is not meant to compete directly with other supervised baselines. Instead, it aims to:
(\textbf{a}) examine the potential origin of the failure in OOD generalization,
(\textbf{b}) provide a rough estimate of the potential improvement achievable with frozen latent representations, and
(\textbf{c}) compare with standard self-supervised baselines and identify the optimal learning algorithms, e.g. SimCLR \citep{chen2020simple}, for training the main encoder.

\textbf{Semi-supervised learning.} We further compare our method directly to other supervised debiasing methods in the context of semi-supervised learning. Here we assume that the training dataset includes only a small amount of labeled data combined with a large amount of unlabeled data. As in linear evaluation, we train a linear classifier on top of the biased encoder by using labeled samples. After obtaining an error set \(E\) of misclassified samples, we finetuned the whole main model by upweighting the identified samples in \(E\) with \(\lambda_{up}\). Note that supervised baselines are restricted to using only a small fraction of labeled samples, while the proposed approach benefits from the abundant unlabeled samples during pre-training of the biased encoder (Pseudo-code in the Appendix section \ref{sec: pseudocode}).

\vspace{-0.2cm}

\section{Results}
\label{sec: results}

\begin{table*}[t]
\caption{{(Supervised learning) Bias-conflict and unbiased accuracy ($\%$) on MIMIC-CXR + NIH. Each \cmark marker represents whether the model requires information on dataset bias. Bias ratio=10\(\%\).}} 
\label{4. table: supervised}
\centering
\resizebox{0.99\textwidth}{!}{
\begin{tabular}{c c c c c c c c c c}
\toprule
\multirow{2}{*}{Accuracy} & {LNL} & {EnD} & {LfF} & {JTT} & {CVaR DRO} & {ERM} & {SimCLR} & {\textbf{Ours}} \\
{} & {\cmark } & {\cmark } & {\xmark} & {\xmark} & {\xmark} & {\xmark} & {\xmark} & {\xmark} \\ 
\midrule
{Conflict} & {43.8$_{\pm0.5}$} & {50.4$_{\pm2.3}$} & {25.2$_{\pm2.1}$} & {47.9$_{\pm0.2}$} & {44.6$_{\pm0.5}$} & {41.7$_{\pm1.2}$} & {35.5$_{\pm1.3}$} & {\textbf{56.8}$_{\pm1.7}$} \\
\midrule
{Unbiased} & {68.1$_{\pm1.0}$} & {\textbf{71.8}$_{\pm1.4}$} & {60.8$_{\pm0.2}$} & {68.9$_{\pm1.0}$} & {65.8$_{\pm1.2}$} & {67.8$_{\pm1.0}$} & {62.0$_{\pm1.4}$} & {69.8$_{\pm0.2}$} \\
\bottomrule
\end{tabular}
}
\end{table*}

\begin{table*}[t]
\caption{(Linear evaluation) Bias-conflict and unbiased test accuracy ($\%$) evaluated on UTKFace and CelebA. Models requiring information on target class or dataset bias in the (pre)training stage are denoted with \cmark in columns Y and B, respectively. Our results are marked in bold to highlight the improvements compared to the mainly interested self-supervised learning baselines (Gray rows).}  
\label{4. table: linear evaluation}
\centering
\begin{tabular}{c c c c c c c c c}
\toprule
\multicolumn{1}{c}{\multirow{2}{*}{Model}} & \multicolumn{1}{c}{\multirow{2}{*}{Y}} & \multicolumn{1}{c}{\multirow{2}{*}{B}} & \multicolumn{2}{c}{UTKFace (age)} & \multicolumn{2}{c}{UTKFace (gender)} & \multicolumn{2}{c}{CelebA (makeup)} \\
\cmidrule{4-9}
\multicolumn{1}{c}{} & \multicolumn{1}{c}{} & \multicolumn{1}{c}{} & Conflict & Unbiased  & Conflict & Unbiased & Conflict & Unbiased \\
\midrule
{LNL} & {\cmark} & {\cmark} & {45.8$_{\pm0.6}$} & {72.6$_{\pm0.3}$} & {73.1$_{\pm1.6}$} & {84.9$_{\pm0.8}$}	& {55.9$_{\pm2.1}$} & {76.0$_{\pm0.6}$} \\
{EnD} & {\cmark} & {\cmark} & {45.3$_{\pm0.9}$} & {72.2$_{\pm0.2}$} & {75.5$_{\pm1.1}$} & {85.5$_{\pm0.4}$}	& {57.3$_{\pm2.4}$} & {76.4$_{\pm1.4}$} \\
\midrule
{JTT} & {\cmark} & {\xmark} & {63.8$_{\pm0.9}$} & {69.4$_{\pm1.3}$} & {71.2$_{\pm0.3}$} & {77.6$_{\pm0.4}$}	& {62.4$_{\pm1.2}$} & {74.7$_{\pm0.8}$} \\
{CVaR DRO} & {\cmark} & {\xmark} & {45.7$_{\pm2.0}$} & {71.4$_{\pm0.3}$} & {68.6$_{\pm1.0}$} & {81.0$_{\pm0.8}$} & {58.0$_{\pm1.7}$} & {76.5$_{\pm0.6}$} \\
\midrule
{ERM} & {\cmark} & {\xmark} & {45.4$_{\pm2.1}$} & {71.0$_{\pm1.2}$} & {65.7$_{\pm1.4}$} & {79.5$_{\pm0.6}$}	& {54.2$_{\pm0.2}$} & {74.1$_{\pm1.4}$} \\
\rowcolor{Gray}
{SimSiam} & {\xmark} & {\xmark} & {28.2$_{\pm0.9}$} & {62.6$_{\pm0.7}$} & {48.5$_{\pm1.0}$} & {69.8$_{\pm0.7}$} & {39.9$_{\pm0.6}$} & {66.7$_{\pm0.6}$} \\
\rowcolor{Gray}
{VICReg} & {\xmark} & {\xmark} & {32.3$_{\pm0.6}$} & {64.6$_{\pm0.3}$} & {51.0$_{\pm1.4}$} & {71.3$_{\pm0.7}$} & {48.6$_{\pm0.6}$} & {71.9$_{\pm0.2}$} \\
\rowcolor{Gray}
{SimCLR} & {\xmark} & {\xmark} & {36.4$_{\pm1.5}$} & {66.3$_{\pm0.6}$} & {56.3$_{\pm0.2}$} & {74.2$_{\pm0.2}$} & {46.9$_{\pm1.0}$} & {69.8$_{\pm0.4}$} \\
\rowcolor{Gray}
{\textbf{DeFund}} & {\xmark} & {\xmark} & {\textbf{59.5}$_{\pm0.8}$} & {\textbf{70.6}$_{\pm0.8}$} & {\textbf{63.7}$_{\pm2.0}$} & {\textbf{74.9}$_{\pm0.9}$} & {\textbf{58.4}$_{\pm0.6}$} & {\textbf{73.1}$_{\pm1.0}$} \\
\bottomrule
\end{tabular}
\end{table*}

\begin{table*}[t]
\caption{(Semi-supervised learning) Accuracy results ($\%$) on CelebA. Label fraction$=10\%$.} 
\label{4. table: semi-supervised}
\centering
\resizebox{0.99\textwidth}{!}{
\begin{tabular}{c c c c c c c c c c}
\toprule
\multirow{2}{*}{Accuracy} & \multicolumn{7}{c}{CelebA (Makeup)} & \multicolumn{2}{c}{CelebA (Blonde)} \\ 
\cmidrule{2-10}
{} & {LNL} & {EnD} & {JTT} & {CVaR DRO} & {ERM} & {SimCLR} & {\textbf{DeFund}} & {JTT} & {\textbf{DeFund}} \\
\midrule
{Conflict} & {55.7$_{\pm1.4}$} & {55.3$_{\pm1.5}$} & {51.5$_{\pm1.9}$} & {55.6$_{\pm1.5}$} & {51.5$_{\pm1.1}$} & {50.5$_{\pm4.7}$} & {\textbf{60.5}$_{\pm0.4}$} & {70.6$_{\pm1.0}$} & {\textbf{75.1}$_{\pm0.8}$} \\
\midrule
{Unbiased} & {75.6$_{\pm0.5}$} & {\textbf{76.2}$_{\pm0.8}$} & {71.4$_{\pm1.3}$} & {75.7$_{\pm1.0}$} & {73.1$_{\pm0.3}$} & {71.6$_{\pm1.9}$} & {75.6$_{\pm0.2}$} & {78.8$_{\pm1.7}$} & {\textbf{85.8$_{\pm0.3}$}} \\
\bottomrule
\end{tabular}
}
\end{table*}

\vspace{-0.2cm}
\subsection{Methods}
{\textbf{Dataset.} We evaluate several supervised and self-supervised baselines on \textbf{MIMIC-CXR + NIH} \citep{li2023partition}, \textbf{UTKFace} \citep{zhang2017age} and \textbf{CelebA} \citep{liu2015deep} in which prior work reported poor generalization performance due to spurious correlations (Dataset details in Appendix).}

For MIMIC-CXR + NIH, we mixed the MIMIC-CXR \citep{johnson2019mimic} and NIH \citep{wang2017chestx} following \cite{li2023partition} where the target categories are \texttt{no finding} and \texttt{pneumonia}. Most \texttt{pneumonia} images are collected from MIMIC-CXR, while most \texttt{no finding} images are from NIH. In other words, the biases come from systematic differences in data sources, where the classifier may erroneously rely on spurious radiographic features tied to variations in data acquisition pipelines \citep{degrave2021ai} instead of true pathological indicators (Example images in Figure \ref{supple fig: examples}).

For UTKFace, we conduct binary classifications using  (\texttt{Gender}, \texttt{Age}) and (\texttt{Race}, \texttt{Gender}) as (target, spurious) attribute pair, which we refer to UTKFace (age) and UTKFace (gender), respectively. For CelebA, we consider (\texttt{HeavyMakeup}, \texttt{Male}) and (\texttt{Blonde Hair}, \texttt{Male}) as (target, spurious) attribute pairs, which are referred to CelebA (makeup) and CelebA (blonde), respectively. The results of CelebA (blonde) are reported in Appendix \ref{sec: CelebA (blonde)}. Following \cite{nam2020learning, hong2021unbiased}, we report bias-conflict accuracy together with unbiased accuracy, which is evaluated on the explicitly constructed validation set. We exclude the dataset in Figure \ref{2 fig: rank reg} based on the observations that the SimCLR models are already invariant w.r.t spurious attributes.  

\textbf{Baselines.} We mainly target baselines consisting of recent advanced self-supervised learning methods, SimCLR \citep{chen2020simple}, VICReg \citep{bardes2021vicreg}, and SimSiam \citep{chen2021exploring}, which can be categorized into contrastive (SimCLR) and non-contrastive (VICReg, SimSiam) methods. We further report the performance of vanilla networks trained by ERM, and other supervised debiasing methods such as LNL \citep{kim2019learning}, EnD \citep{tartaglione2021end}, and upweighting-based algorithms, JTT \citep{liu2021just} and CVaR DRO \citep{levy2020large}, which can be categorized into methods that leverage annotations on dataset bias (LNL, EnD) or not (JTT, CVaR DRO). 
%JTT is a failure-based debiasing approach which upweights the bias-conflicting samples identified by model trained with capacity control techniques, e.g., early-stopping, strong \(\ell_2\) regularization. 
%CVaR DRO models a wide range of potential distribution shifts, and minimize the specific type of worst-case loss over all subsets of the training set of a certain size \citep{duchi2019distributionally}.

\textbf{Optimization setting.} Both bias and main encoder is pretrained with SimCLR \citep{chen2020simple}  for 100 epochs on UTKFace, and 20 epochs on CelebA, respectively, using ResNet-18, Adam optimizer and cosine annealing learning rate scheduling \citep{loshchilov2016sgdr}. We use a MLP with one hidden layer for projection networks as in SimCLR. All the other baseline results are reproduced by tuning the hyperparameters and optimization settings using the same backbone architecture. We report the results of the model with the highest bias-conflicting test accuracy over those with improved unbiased test accuracy compared to the corresponding baseline algorithms, i.e., SimCLR for ours (More experimental details in Appendix \ref{sec: experimental setup}).
%The same criteria are applied to supervised baselines, while JTT often sacrifices unbiased accuracy for highly improved bias-conflict accuracy.   
%Though we observe that some baselines may cost overall performance degradation for improved bias-conflicting test accuracy due to the tradeoff reported in \citet{sagawa2019distributionally}. Details about other simulation settings, datasets and baselines are provided in the supplementary material. 

\subsection{Evaluation results}

\textbf{Supervised learning.} To quantify the effectiveness of the rank regularization in-depth, we first consider a standard supervised debiasing scenario as similarly done in Table \ref{2. table: MultiCMNIST}. For a MIMIC-CXR + NIH dataset, we found that the proposed framework outperforms other supervised baselines with respect to bias-conflict accuracy. Table \ref{supp table: mimic} in the Appendix shows that the rank-regularized networks effectively discover the bias-conflicting samples which are consistent with Table \ref{2 table: precision}, \ref{2 table: precision_jtt}, and \ref{2. table: MultiCMNIST}.

%\vspace{-0.4cm}
\textbf{Linear evaluation.} We also found that DeFund outperforms every self-supervised baseline by a large margin in a linear evaluation protocol, including SimCLR, SimSiam and VICReg, with respect to both bias-conflict and unbiased accuracy (Table \ref{4. table: linear evaluation}). Moreover, in some cases, DeFund even outperforms ERM models or supervised debiasing approaches regarding bias-conflict accuracy. Note that there is an inherent gap between ERM models and self-supervised baselines, roughly \(8.7\%\) on average. Moreover, we found that non-contrastive learning methods generally perform worse than the contrastive learning method. This warns us against training the main model using a non-contrastive learning approach, while it may be a viable option for the biased model. Results of the proposed framework with non-contrastive learning methods are provided in the Appendix section \ref{sec: non_CL}. 

\textbf{Semi-supervised learning.} To compare the performance of supervised and self-supervised methods in a more practical and fair scenario, we sample \(10\%\) of the labeled CelebA training dataset at random for each run. The remaining \(90\%\) samples are treated as unlabeled ones and engaged only in pretraining encoders for self-supervised baselines. Labeled samples are provided equally to both supervised and self-supervised methods. 

Remarkably, Table \ref{4. table: semi-supervised} and Table \ref{supple table: blonde semi supervised} in Appendix show that the proposed framework outperforms other state-of-the-art supervised debiasing methods. Existing upweighting protocols, such as JTT, fail to prevent deep networks from memorizing minority counterexamples. However, the proposed framework can fully utilize unlabeled samples with contrastive learning to prevent memorization. Existing bias-conflicting sample mining algorithms may be affected by the implicit bias of overparameterized networks, but this is unlikely to happen with the proposed framework since it only trains a simple linear classifier on top of a frozen biased encoder to identify such samples.

\begin{table}[htbp]
\centering
    \begin{subtable}[h]{0.49\textwidth}
        \centering
        \resizebox{0.99\textwidth}{!}{
            \begin{tabular}{c c c c c c c}
            \toprule
            \multicolumn{1}{c}{\multirow{2}{*}{Method}} & \multicolumn{2}{c}{UTKFace (age)} & \multicolumn{2}{c}{UTKFace (gender)} & \multicolumn{2}{c}{CelebA (makeup)} \\
            \cmidrule{2-7}
            \multicolumn{1}{c}{} & Conflict & Unbiased & Conflict & Unbiased & Conflict & Unbiased \\
            \midrule
            \rowcolor{Gray}
            {SimCLR} & {36.4} & {66.3} & {56.3} & {74.2} & {46.9} & {69.8} \\
            {+ Rank reg} & {26.6} & {61.3} & {50.9} & {70.3} & {43.9} & {68.3} \\
            {+ Upweight} & {53.0} & {64.6} & {58.3} & {74.5} & {50.1} & {70.4} \\
            \rowcolor{Gray}
            {\textbf{DeFund}} & {\textbf{59.5}} & {\textbf{70.6}} & {\textbf{63.7}} & {\textbf{74.9}} & {\textbf{58.4}} & {\textbf{73.1}} \\
            \bottomrule
            \end{tabular}
            }
       \caption{Ablation study}
       \label{4. table: ablation}
    \end{subtable}
    \begin{subtable}[h]{0.49\textwidth}
        \centering
        \resizebox{0.99\textwidth}{!}{
            \begin{tabular}{c c c c c c c}
            \toprule
            \multicolumn{1}{c}{\multirow{2}{*}{Method}} & \multicolumn{2}{c}{UTKFace (age)} & \multicolumn{2}{c}{UTKFace (gender)} & \multicolumn{2}{c}{CelebA (makeup)} \\
            \cmidrule{2-7}
            \multicolumn{1}{c}{} & Precision & Recall & Precision & Recall & Precision & Recall \\
            \midrule
            {SimCLR} & {68.31} & {44.63} & {\textbf{33.36}} & {39.59} & {52.25} & {28.23} \\
            \midrule
            {\textbf{DeFund}} & {\textbf{68.67}} & {\textbf{75.94}} & {29.98} & {\textbf{50.93}} & {\textbf{55.29}} & {\textbf{32.46}} \\
            \bottomrule
            \end{tabular}
            }
       \caption{Precision and recall} 
       \label{4. table: precision}
    \end{subtable}
    \caption{(\textbf{a}) Ablation study on introduced modules. (\textbf{b}) Precision and recall ($\%$) of bias-conflicting samples identified by SimCLR and our biased model. Both case used linear evaluation. }
\end{table}

%\vspace{-0.5cm}
\textbf{Ablation study.} To quantify the extent of performance improvement achieved by each introduced module, we compared the linear evaluation results of (\textbf{a}) vanilla SimCLR, (\textbf{b}) SimCLR with rank regularization, (\textbf{c}) SimCLR with upweighting error set \(E\) of the main model, and (\textbf{d}) DeFund. Note that (\textbf{c}) does not use a biased model at all. Table \ref{4. table: ablation} shows that every module plays an important role in OOD generalization. Considering that the main model is already biased to some extent, we found that bias-conflict accuracy can be improved even without a biased model, where the error set \(E\) of the biased model further boosts the generalization performance. We also measures the precision and recall of identified bias-conflicting samples in E, finding that the biased model detects more diverse bias-conflicting samples than the baseline (Table \ref{4. table: precision}). The improvement of recall in CelebA may seem marginal, but it is significant given the larger number of samples compared to UTKFace.

\textbf{Computational costs.} The proposed framework is computationally affordable as it only trains the linaer classifier (linear eval.) or finetune networks with a few epochs, e.g., about 30 epochs for UTKFace in debiasing stage. Self-supervised pre-training and linear evaluation takes 19.3 and 4.5 minutes with a single NVIDIA GeForce RTX 2080Ti, respectively.

\vspace{-0.3cm}
\section{Conclusion}
\textbf{Contributions.} We present a novel solution to the challenging self-supervised debiasing, an important problem that has received little attention so far. Specifically, we (\textbf{a}) unveil the inductive bias towards encoding low effective rank representations in the presence of spurious correlations. Based on these findings, we (\textbf{b}) design a rank regularization that amplifies the feature redundancy by reducing the spectral entropy of latent representations. Then we (\textbf{c}) design a debiasing framework empowered by the biased model pretrained with abundant unlabeled samples.

\section{Ethics and reproducibility statements}
\textbf{Ethics statement.} While this work has focused on encoding biased representations, more advances should be made in learning both biased and debiased representations. We found that explicit decorrelation of feature components in SimCLR does not lead to debiased representations. Moreover, one potential negative impact of the proposed framework could be the perpetuation of biases in data that are already present in society. The self-supervised biased encoder may amplify existing biases in the data, which could lead to further discrimination of certain populations. This must be prevented through proper regulation.

\textbf{Reproducibility statements.} We upload a file containing the code for our main experiments as a supplementary material. Furthermore, to ensure maximum reproducibility, pseudo-codes and hyper-parameter configurations are described in the Appendix \ref{sec: pseudocode} and \ref{sec: experimental setup}, respectively. Rest assured, we will fully open-source our code and pretrained models to reproduce all experiments in the paper.

\bibliography{iclr2024_conference}
\bibliographystyle{iclr2024_conference}

\newpage
\appendix

\noindent\makebox[\textwidth][c]{\Large\bfseries Appendix}

The supplementary material is organized as follows. We begin with providing the algorithm of DeFund, followed by more discussions on the related works. Then we provide additional results and analyses in section \ref{sec: additional}. Optimization setting, hyperparameter configuration and other experimental details are provided in section \ref{sec: experimental setup}.

\section{Pseudocode}
\label{sec: pseudocode}
We provide the pseudo-code of the supervised and self-supervised version of the proposed debiasing framework here due to the limited space. The only minor difference between these two versions simply lies in the choice of training methodology, specifically whether the main and biased models are trained using supervised or self-supervised learning algorithms, respectively. Table \ref{2. table: MultiCMNIST} and \ref{4. table: supervised} are based on Algorithm 2, while Table \ref{4. table: linear evaluation}, \ref{4. table: semi-supervised}, \ref{4. table: ablation} and \ref{4. table: precision} are based on Algorithm 1.

\begin{algorithm}
\label{3. algorithm}
	\caption{Debiasing Framework with unlabeled data (DeFund, self-supervised learning)} 
	\begin{algorithmic}[1]
	    \STATE {\bfseries Input:} \(D_l=\{(x_k, y_k)\}_{k=1}^{N_1}\), \(D_u=\{x_k\}_{k=1}^{N_2}\) for semi-supervised learning (\(N_2 \gg N_1 \)), or \(\varnothing\) for linear evaluation, \(D = D_l \cup D_u\), batch size \(n\), structure of \(f^{bias}\) and \(f^{main}\). 
	    \STATE
        
        \STATE \textbf{Stage 1.} \textit{Pretraining encoders}
            \FOR {subsampled minibatch \(X=\{ x_k \}_{k=1}^n\) from D}
                \STATE Update \(\theta\) of \(f^{bias}_\theta\) with SimCLR NT-Xent loss and \(\lambda_{reg} \ell_{reg}(X; \theta)\).
                \STATE Update \(\phi\) of \(f^{main}_\phi\) with SimCLR NT-Xent loss.
            \ENDFOR
            
        \STATE Obtain pretrained parameters \(\hat{\theta}\) and \(\hat{\phi}\). 
        \STATE
        
        \STATE \textbf{Stage 2.} \textit{Downstream task}
            \STATE Freeze \(f^{bias}_{\hat{\theta}}\) and train \(f^{cls}_{W_b}\) with \(D_l\). Identify the error set \(E \subset D_l\) with trained \(f^{bias}\).
            
            \IF {Linear evaluation} 
                \STATE Freeze \(f^{main}_{\hat{\phi}}\) and train \(f^{cls}_{W_m}\) with \(\ell_{debias}(D_l; W_m)\)
            \ELSIF {Semi-supervised learning}
                \STATE Finetune \(f^{main}\) with \(\ell_{debias}(D_l; W_m, \phi)\) where $\phi$ is initialized with $\hat{\phi}$.
            \ENDIF
	\end{algorithmic} 
\end{algorithm}

\begin{algorithm}
\label{3. algorithm_sup}
	\caption{Debiasing Framework with rank regularization (DeRank, supervised learning)} 
	\begin{algorithmic}[1]
	    \STATE {\bfseries Input:} \(D=\{(x_k, y_k)\}_{k=1}^{N}\), batch size \(n\), structure of \(f^{bias}\) and \(f^{main}\). 
	    \STATE
        
        \STATE \textbf{Stage 1.} \textit{Training biased model}
            \FOR {subsampled minibatch \(X=\{x_k\}_{k=1}^n, Y=\{y_k\}_{k=1}^n\) from D}
                \STATE Update \(\theta, W_b\) of \(f^{bias}\) with standard cross entropy loss and \(\lambda_{reg} \ell_{reg}(X; \theta)\).
            \ENDFOR
            
        \STATE Obtain pretrained parameters \(\hat{\theta}\) and \(\hat{W}_b\).
        \STATE Identify the error set \(E \subset D\) with trained \(f^{bias}\).
        \STATE
        
        \STATE \textbf{Stage 2.} \textit{Training main model}
        \STATE Train \(f^{main}\) with \(\ell_{debias}(D; W_m, \phi)\).
            
	\end{algorithmic} 
\end{algorithm}

\section{More related works}
\label{sec: related works}

\textbf{Learning debiased representations.} \citet{robinson2021can} proposes an opposite-directional approach compared to our framework to improve generalizations of self-supervised learning. It aims to overcome the feature suppression and learn a wide variety of features by Implicit Feature Modification (IFM), which adversarially perturbs feature components of the current representations used to discriminate instances, thereby encouraging the encoder to use other informative features. We observed that IFM improves the bias-conflict accuracy by about 1$\%$ on UTKFace (age) in Table \ref{supp table: ifm}, which is roughly consistent with the performance gains on the standard benchmarks, e.g., STL10, reported in the original paper. However, its performance gain is relatively marginal compared to the proposed framework. 

\begin{table}[htbp]
\caption{Results of Implicit Feature Modification \citep{robinson2021can} with SimCLR on UTKFace (age). we denote $\epsilon$ as the adversarial budget of feature modification as in the original paper.} 
\label{supp table: ifm}
\centering
\begin{tabular}{c c c c c}
\toprule
{Accuracy} & {SimCLR} & {$\epsilon=0.05$} & {$\epsilon=0.1$} & {$\epsilon=0.5$} \\
\midrule
{Bias-conflict ($\%$)} & {36.4} & {\textbf{37.5}} & {36.4} & {33.7} \\
\midrule
{Unbiased ($\%$)} & {66.3} & {\textbf{66.5}} & {66.2} & {64.6} \\
\bottomrule
\end{tabular}
\end{table}

\textbf{Discovering bias without supervision.}
In practice, several limitations exist against gleaning more labeled samples: labeling budget, expert-level knowledge required for labeling, data privacy, etc. In this regard, most training samples lack annotations on the spuriously correlated attributes.

To mitigate these problems, several works aim to discover biases without bias annotations. \citet{liu2021just} reveals that the standard ERM model may serve as a bias-capturing model if one trains it with strong capacity control. \citet{yaghoobzadeh2019increasing} shows that forgettables, or examples that have been forgotten at least once, contain more minority examples, and proposes a novel robust learning framework by fully exploiting the identified forgettable examples. \citet{li2021discover} obtains a biased attribute hyperplane of the generative models, which can help identify semantic biases by generating bias-traversal images. \citet{li2022discover} introduces the discoverer model, which uncovers multiple unknown biases such that the difference of averaged predicted probabilities on the target attribute in two groups is maximized. \citet{lang2021explaining} proposes a novel framework, StylEx, which trains a styleGAN to specifically visualize multiple attributes underlying the classifier decisions. 

While substantial advances have been made in discovering the unknown biases of neural networks without bias labels, these works still inevitably require target labels. In contrast, we consider a very challenging scenario that has received little attention so far: self-supervised debiasing. In this regard, our work addresses the following open problems/questions: 
\begin{itemize}
    \item Can we learn biased/debiased representations by using unlabeled samples?
    \item What is the fundamental difference between biased and debiased representations?
    \item Is supervised debiasing robust despite decreasing the number of labeled samples?
    \item How can bias-conflicting samples be discovered by leveraging information from unlabeled samples? 
    \item Many recent works have reported the limitations of self-supervised learning (SSL) in OOD generalization. How can we overcome such limitations?
\end{itemize}

\textbf{Mitigating bias with reweighting.} Recently, \cite{kirichenko2022last} have reported an intriguing observation: Simple last layer retraining, so-called Deep Feature Reweighting (DFR), can match or outperform state-of-the-art approaches on spurious correlation benchmarks.  \citet{kirichenko2022last} shows that biased classifiers still often learn core features associated with the desired attributes of the data. Based on these observations, they probe invariant features for the reweighting by leveraging explicit group-balanced dataset $\hat{D}$. 

We compare the proposed framework with DFR as follows. First, while DFR and the proposed framework can mitigate the bias in representations by retraining the last linear layer, our method is not restricted to such last-layer retraining. Instead, the semi-supervised learning scenario is a more practical application of the proposed method. Specifically, we can fine-tune representations by fully exploiting both unlabeled and labeled samples, which improves the performance compared to the last layer retraining in Table \ref{4. table: semi-supervised}. In contrast, DFR trains a linear classifier while freezing the pretrained representations as-is. More importantly, DFR requires pretrained networks or fully labeled datasets where we consider a more challenging scenario without such assumptions. Moreover, DFR does not use mining bias-conflicting samples in the training set. Specifically, DFR trains a new classification head from scratch on the available group-balanced data $\hat{D}$. In \cite{kirichenko2022last}, the reweighting dataset $\hat{D}$ often consists of a random group-balanced subset of the training or validation data. In other words, DFR is not designed to identify the bias-conflicting samples but exploits the existing group annotations. Considering practical situation with several limitations against collecting more labeled samples, it remains unclear how to obtain the group-balanced dataset $\hat{D}$ with sufficient number of samples in the absence of prior information on the dataset bias. In contrast, the proposed framework can leverage the explicit set $\hat{D}$ if accessible, \textit{as well as} identifying the unknown bias-conflicting samples in the training set.

\vspace{-0.4cm}

\section{Additional results}
\label{sec: additional}
Our additional results can be roughly categorized into: (1) more observations related to the rank reduction, (2) rank regularization in self-supervised learning, and (3) an examination of the potential of existing hyperparameters as a bias controller. Our observations include the rank reduction trends in CIFAR-10C and Vision Transformer (ViT, \cite{dosovitskiy2020image}), followed by rank regularization results with a moderate level of bias, and results of nuclear norm regularization. Then we present a simple synthetic simulation on the behavior of rank-regularized encoder. Then the potential of using shallow networks as the bias-capturing model will be discussed, followed by additional results on non-contrastive methods, MIMIC-CXR + NIH, and CelebA (blonde). Lastly, we provide additional analysis on relations between existing hyperparameters of self-supervised learning and effective rank. 

\vspace{-0.2cm}

\subsection{More observations}
\label{sec: more observations}

\begin{figure}[htbp]
\centering
\begin{subfigure}[c]{0.3\textwidth}
\includegraphics[width=\textwidth]{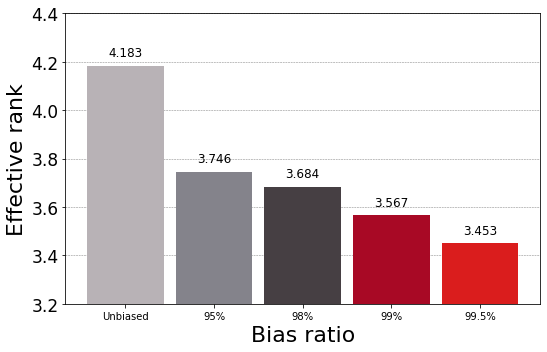} 
\subcaption[c]{CIFAR-10C}
\label{supple fig: rank_cifar}
\end{subfigure}
\begin{subfigure}[c]{0.3\textwidth}
\includegraphics[width=\textwidth]{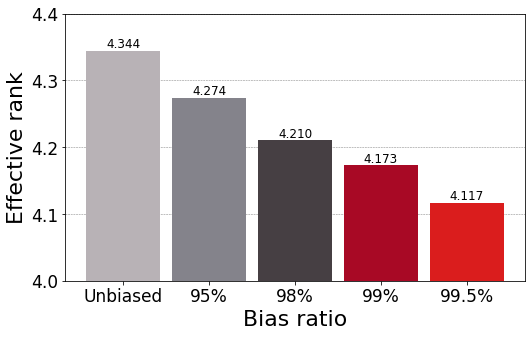} 
\subcaption[c]{CMNIST (ViT)}
\label{supple fig: rank_cmnist_vit}
\end{subfigure}
\begin{subfigure}[c]{0.3\textwidth}
\includegraphics[width=\textwidth]{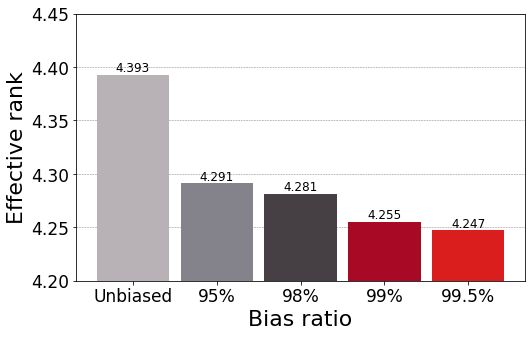} 
\subcaption[c]{CIFAR-10C (ViT)}
\label{supple fig: rank_cifar_vit}
\end{subfigure}
\caption{Effective rank measured with (\textbf{a}) CIFAR-10C (ResNet-18), (\textbf{b}) CMNIST (ViT) and (\textbf{c}) CIFAR-10C (ViT).
\vspace{-0.2cm}
}
\label{supple fig: rank}
\end{figure}

\textbf{Rank reduction.} Figure \ref{supple fig: rank_cifar} shows that the rank of latent representations from a penultimate layer of classifier decreases as the bias ratio increases in CIFAR-10C. In Table \ref{supple table: cmnist and cifar10c}, we supplement the unbiased test accuracy of CMNIST and CIFAR-10C from the experiments presented in Figure \ref{2 fig: cmnist rank} and \ref{supple fig: rank_cifar}, respectively. Moreover, similar rank reduction trends are observed in Vision Transformer (ViT, \cite{dosovitskiy2020image}). We train ViT on CMNIST and CIFAR-10C for 2000 and 10000 iterations, respectively, with Adam optimizer of learning rate 0.001, patch size 4, dimension of output tensor 128, number of transformer blocks 6, number of heads in multi-head Attention layer 4, dropout rate 0.2 and dimension of the MLP (FeedForward) layer 1024. Figure \ref{supple fig: rank_cmnist_vit}, \ref{supple fig: rank_cifar_vit} show that the effective rank of the output of the Transformer encoder $\textbf{z}_L^0$ (notation follows the original paper) decreases as bias ratio increases. 

\begin{table}[hbtp]
\caption{Unbiased test accuracy ($\%$) on CMNIST and CIFAR-10C measured with varying bias ratio $r$. The model trained with unbiased dataset ($r=10\%$) serves as a baseline.}
\label{supple table: cmnist and cifar10c}
\centering
\begin{tabular}{c c c c c c}
\toprule
{Dataset} & {Unbiased} & {$r=95\%$} & {$r=98\%$} & {$r=99\%$} & {$r=99.5\%$} \\
\midrule
{CMNIST} & {99.87} & {88.27} & {68.13} & {36.21} & {13.61} \\
\midrule
{CIFAR-10C} & {78.71} & {46.15} & {34.18} & {26.76} & {20.94} \\
\bottomrule
\end{tabular}
\end{table}

\begin{table}[htbp]
\caption{Ablation study of rank regularization on weakly biased CMNIST (Bias ratio=60$\%$). Our rank-regularized model is trained with $\lambda_{reg}=50$. For a fair comparison, all the other experimental settings are fixed. Bias-aligned accuracy, bias-conflict accuracy, precision and recall of identified bias-conflicting samples are reported.}
\label{supple table: weak bias}
\centering
\begin{tabular}{c c c c c}
\toprule
{Methods} & {Align ($\%$)} & {Conflict ($\%$)} & {Precision ($\%$)} & {Recall ($\%$)} \\
\midrule
{ERM} & {99.49} & {97.81} & {79.55} & {0.87} \\
\midrule
{Ours} & {96.25} & {38.15} & {91.56} & {60.97} \\
\bottomrule
\end{tabular}
\end{table}

\textbf{Rank regularization with moderate level of bias.} To study the compatibility of rank regularization with weak spurious correlations, we apply the rank regularization to the moderately biased CMNIST, i.e., bias ratio=60$\%$. Table \ref{supple table: weak bias} shows that the rank regularization works well in this natural setting. This implies that the rank regularization can be leveraged to reveal the moderate level of bias embedded in the representations, which is supported by the empirical results of other general datasets, e.g., Waterbirds, UTKFace or CelebA.

\textbf{Nuclear norm regularization.} While the proposed rank regularization controls the auto-correlation matrix inspired from \ref{2 fig: unbias corr}, one may regularize nuclear norm of the latent representations, which is a convex relaxation of a matrix rank. 

To compare the quality of biased representations, we call $\text{DeFund}_{nu}$ as the proposed debiasing framework with normalized nuclear norm regularization, instead of \eqref{2.2 eq: rank reg}. Specifically, for a normalized nuclear norm, the absolute singular values are summed and then divided with the feature dimension. From our preliminary analysis in Table \ref{supple table: nuclear norm} below, the performance of nuclear norm regularization was underperformed by the proposed rank regularization in \eqref{2.2 eq: rank reg}. 
Moreover, for the case of the nuclear norm, top singular values are significantly large, as shown in Figure \ref{2 fig: lambda}, so that the distributional property of singular values may be obfuscated in the nuclear norm as shown in Table \ref{supple table: nuclear norm analysis}. This suggests that while nuclear norm may be a candidate for rank regularizer with a solid theoretical background, we recommend using the effective rank in feature analysis.

\begin{table}[htbp]
\caption{(Linear evaluation) Bias-conflict and unbiased test accuracy ($\%$) evaluated on UTKFace and CelebA. $\text{DeFund}_{nu}$ refers to the proposed framework with nuclear norm regularization.}
    \label{supple table: nuclear norm}
    \centering
    \begin{tabular}{c c c c c}
    \toprule
        \multicolumn{1}{c}{\multirow{2}{*}{Model}} & \multicolumn{2}{c}{UTKFace (age)} & \multicolumn{2}{c}{CelebA (makeup)} \\
        \cmidrule{2-5}
        {} & {Conflict} & {Unbiased} & {Conflict} & {Unbiased} \\
        \midrule
         $\text{DeFund}_{nu}$ & $53.9_{\pm 0.3}$ & $67.5_{\pm 0.3}$ & $52.1_{\pm 0.5}$ & $72.5_{\pm 0.1}$ \\
         \midrule
         {\textbf{DeFund}} & {\textbf{59.5}$_{\pm0.8}$} & {\textbf{70.6}$_{\pm0.8}$} & {\textbf{58.4}$_{\pm0.6}$} & {\textbf{73.1}$_{\pm1.0}$} \\
         \bottomrule
    \end{tabular}
\end{table}

\begin{table}[htbp]
\caption{Normalized nuclear norm (norm / dimension) measured in CMNIST and CIFAR-10C with varying bias ratios.}
\label{supple table: nuclear norm analysis}
\centering
\begin{tabular}{c c c c c c}
\toprule
{Dataset} & {Unbiased} & {95($\%$)} & {98($\%$)} & {99($\%$)} & {99.5($\%$)} \\
\midrule
{CMNIST} & {2.47} & {2.56} & {2.56} & {2.59} & {2.46} \\
\midrule
{CIFAR-10C} & {7.12} & {5.92} & {6.34} & {6.54} & {6.51} \\
\bottomrule
\end{tabular}
\end{table}

\begin{figure}[!b]
\centering
\begin{subfigure}[c]{0.67\textwidth}
\includegraphics[width=\textwidth]{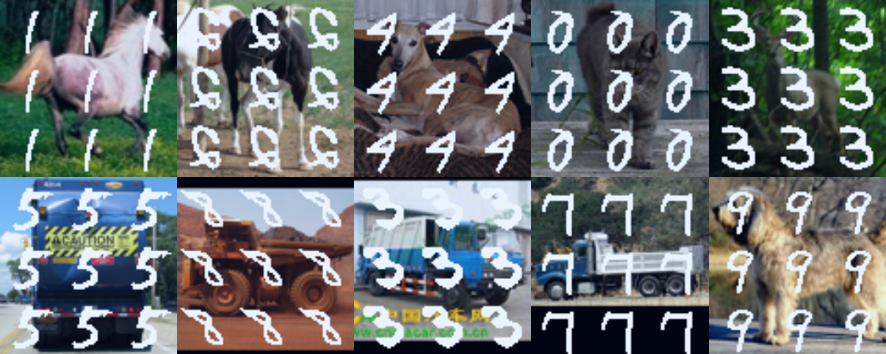} 
\subcaption[c]{Sample images}
\label{supple fig: DigitsOnSTL10}
\end{subfigure}
\begin{subfigure}[c]{0.31\textwidth}
\includegraphics[width=\textwidth]{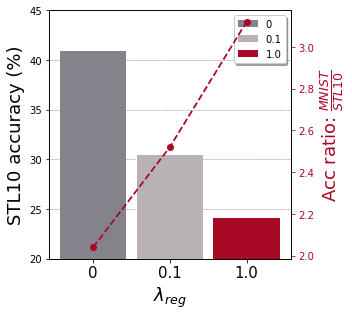} 
\subcaption[c]{Evaluation results}
\label{supple fig: DigitsOnSTL10_eval}
\end{subfigure}
\caption{(\textbf{a}) Sample images from DigitsOnSTL10 dataset. (\textbf{b}) Test accuracy of STL10 classifier and bias metric.}
\end{figure}

\textbf{Behavior of rank-regularized encoder.} Here, we present a simple simulation which conceptually clarifies the impacts of rank regularization in self-supervised learning. Inspired from \cite{chen2020simple, robinson2021can}, we create a DigitsOnSTL10 dataset as in Figure \ref{supple fig: DigitsOnSTL10} where MNIST images are randomly selected and placed on top of the STL10 images. After self-supervised representation learning, we train two independent linear classifiers on top of the freezed representations, where we provide label of foreground MNIST digit for one classifier, and label of background STL10 object class for the other. After training linear classifiers, we measure the ratio of MNIST classifier test accuracy to STL10 classifier test accuracy, which we treat as a proxy of ratio of spuriously correlated features to invariant features, i.e., degree of bias in representations. Intuitively, the proposed bias metric increases as the encoder focus more on the short-cut attribute, i.e., MNIST digit. 

We measure the bias metric on the representations of ResNet-18 encoders trained by SimCLR \citep{chen2020simple} together with rank regularization loss \(\lambda_{reg} \ell_{reg}\), where $\lambda_{reg} > 0$ is a balancing hyperparameter. As denoted in the main paper, we apply regularization not on the output of projection networks but directly on the output of base encoder, which makes it fully agnostic to networks architecture. Figure \ref{supple fig: DigitsOnSTL10_eval} shows that the rank regularization exacerbates the ``feature suppression" phenomenon revealed by \cite{chen2021intriguing}. The representation becomes more biased as it is trained with stronger regularization. While the overall performance of self-supervised learning may be upper-bounded due to the constraint on effective dimensionality \citep{jing2021understanding}, we observe in Figure \ref{supple fig: DigitsOnSTL10_eval} that the bias-conflict accuracy is primarily sacrificed compared to the bias-aligned accuracy. Coupled with results in section \ref{sec: results}, this result implies that rank regularization can amplify bias in self-supervised encoder.

Moreover, we have conducted an additional experiment to better understand the biased downstream application problem. We first train the encoder on unbiased CMNIST using SimCLR. By unbiased, we mean that the background color in the training images is randomly assigned, unlike the images shown in Figure \ref{2 fig: cmnist_aligned}. Subsequently, we train the linear classifier on top of the encoder using (\textbf{a}) Biased CMNIST samples with a bias ratio of 99.5$\%$, and (\textbf{b}) Unbiased CMNIST samples. 

As shown in Table \ref{supple table: biased downstream}, training the linear classifier with unbiased samples (\textbf{b} case) leads to the unbiased model, which works evenly well on every group. Despite training the encoder on a fully unbiased dataset, the use of biased samples in the downstream task results in a significant drop in the bias-conflict test accuracy. These findings highlight the potential risks associated with using biased training samples directly in downstream applications, as biased samples may inadvertently involve spurious factors that are correlated with the bias (such as the background color in this example).

\begin{table}[htbp]
\caption{Linear evaluation results on the CMNIST with varying bias ratio in the downstream dataset.}
    \label{supple table: biased downstream}
    \centering
    \begin{tabular}{c c c}
    \toprule
         Bias ratio & 99.5$\%$ & Unbiased ($10\%$) \\
         \midrule
         {Aligned ($\%$)} & 99.79 & 96.27 \\
         \midrule
         {Conflict ($\%$)} & 69.99 & 96.14 \\
         \bottomrule
    \end{tabular}
\end{table}

\begin{table}[htbp]
\caption{Comparison study on the depth of biased networks. Both networks are trained with target labels on CIFAR-10C (Bias ratio=95$\%$). For UTKFace (age) and CelebA (makeup), both networks are pretrained with SimCLR followed by last linear layer training. Reported in ($\%$).} 
\label{supple table: shallow}
\centering
\begin{tabular}{c c c c c c c}
\toprule
\multicolumn{1}{c}{\multirow{2}{*}{Networks}} & \multicolumn{2}{c}{CIFAR-10C} & \multicolumn{2}{c}{UTKFace (age)} & \multicolumn{2}{c}{CelebA (makeup)} \\
\cmidrule{2-7}
\multicolumn{1}{c}{} & Precision & Recall & Precision & Recall & Precision & Recall \\
\midrule
{Shallow} & {64.73} & {\textbf{59.50}} & {55.68} & {69.98} & {27.49} & {\textbf{33.79}} \\
\midrule
{ResNet-18} & {\textbf{71.39}} & {51.43} & {\textbf{68.67}} & {\textbf{75.94}} & {\textbf{55.29}} & {32.46} \\
\bottomrule
\end{tabular}
\end{table}

\subsection{Shallow network} Considering the inductive bias of neural networks towards encoding low effective rank representations in this paper, one may ask whether the shallow neural networks can easily learn such simple inductive bias and serve as a bias-capturing network. In this regard, we observe some pros and cons of using a shallow network as the bias model throughout experiments. Specifically, we use a simple convolutional network with three convolution layers as a counterpart of ResNet-18, with feature map dimensions of 64, 128 and 256, each followed by a ReLU activation and a batch normalization. 

In the labeled setting, CIFAR-10C in Table \ref{supple table: shallow} shows a tradeoff between precision and recall of the shallow network: The shallow network improves the recall of identified hard samples, i.e., the fraction of the bias-conflicting samples that are identified, because it is robust to the unintended memorization due to their fewer number of hyperparameters. However, it sacrifices the precision, i.e., the fraction of identified samples that are indeed bias-conflicting because its performance on the bias-aligned samples is degraded due to the low expressivity. 

While the shallow network shows promising results with a simple dataset, the tradeoff worsens in the self-supervised setting with a larger dataset. Table \ref{supple table: shallow} shows that the shallow network may suffer from bad precision. It is conventional wisdom that unsupervised learning benefits more from bigger models than its supervised counterpart \citep{chen2020simple}. Considering this, the general performance of shallow networks may deteriorate in a large-scale self-supervised learning scenario. In this case, the identified error set $E$ contains too many false-positive bias-conflicting samples. While one may improve the performance with good care of hyperparameter tuning, e.g., depth of networks, learning rate, etc., it may be more laborious compared to the proposed framework, which has only a few scalar hyperparameters, e.g., $\lambda_{reg}$.

\subsection{Additional results on MIMIC-CXR + NIH}
\label{sec: mimic}
For a MIMIC-CXR + NIH dataset, we report precision and recall of identified bias-conflicting proxies in Table \ref{supp table: mimic}, showing that the proposed rank-regularization improves the minority mining performance.

\begin{table}
\caption{(MIMIC-CXR + NIH) (\textbf{a}), (\textbf{b}): Precision and recall of identified bias-conflicting samples. (\textbf{c}), (\textbf{d}): bias-aligned and bias-conflicting accuracy (\%) of ERM and our rank-regularized model.} 
\label{supp table: mimic}
\centering
\begin{tabular}{c c c c c}
\toprule
{} & {(\textbf{a}) Precision ($\uparrow$)} & {(\textbf{b}) Recall ($\uparrow$)} & {(\textbf{c}) BA ($\uparrow$)} & {(\textbf{d}) BC ($\downarrow$)} \\
\midrule
{ERM} & {52.21} & {54.31} & {95.15} & {29.75} \\
\midrule
{Rank reg.} & {\textbf{55.93}} & {\textbf{70.37}} & {\textbf{96.55}} & {\textbf{17.10}} \\
\bottomrule
\end{tabular}
\end{table}

\subsection{Additional results on CelebA}
\label{sec: CelebA (blonde)}
We report the results of CelebA (blonde) in here due to the limited space. Detailed information on the dataset and simulation settings is provided in the section \ref{sec: experimental setup}. Following \cite{sagawa2019distributionally, liu2021just}, we report worst-group and average accuracy because CelebA (blonde) includes abundant samples in (\texttt{Blonde Hair}=0, \texttt{Male}=0) bias-conflicting group. The number of training samples in each group is provided in Table \ref{supple table: num data}.

Table \ref{supple table: blonde linear eval} shows that DeFund outperforms not only every self-supervised baseline, but also ERM, CVaR DRO, and LfF \cite{nam2020learning} in linear evaluation. Table \ref{supple table: blonde semi supervised} shows that DeFund outperforms all the other baseline methods in semi-supervised learning, which is consistent with Table \ref{4. table: semi-supervised} of the main paper.

Moreover, recent works unveil that CelebA (blonde) exhibits a large class imbalance which in turn correlates with a large
group imbalance. Recent studies \cite{hong2021unbiased, idrissi2022simple} found that both target classes are biased toward a non-Male bias class in CelebA (blonde) which obfuscates whether the dataset is indeed biased. In this regard, \cite{idrissi2022simple} observed that the simple class balancing serves as a powerful baseline due to the class imbalance. This directly motivates us to alleviate the class imbalance and focus on the dataset bias itself. Following \cite{hong2021unbiased}, we randomly subsample images from (\texttt{Blonde Hair}=0, \texttt{Male}=0) group so that two target classes are biased toward different bias classes. The number of training samples before and after subsampling is provided in Table \ref{supple: num celebA (blonde)} and \ref{supp table: subsampled CelebA (blonde)}, respectively. 
Table \ref{supp table: subsampled CelebA acc} shows that DeFund outperforms JTT with respect to both worst-group and average accuracy, where its bias-conflict-accuracy-version is also reported in Table \ref{4. table: semi-supervised} of the main paper. These additional results imply that the proposed framework ensures reliable performance in the presence of strong spurious correlations.

\begin{table}[htbp]
\caption{(Linear evaluation) Worst-group and average accuracy ($\%$) evaluated on CelebA (blonde). Results of ERM, CVaR DRO, LfF \citep{nam2020learning} and JTT are come from Table 1 of the original JTT paper \citep{liu2021just}. Each first and second \cmark marker represents whether the model requires information on target class or dataset bias in pretraining stage, respectively.} 
\label{supple table: blonde linear eval}
\centering
\begin{tabular}{c c c c c c c c c c}
\toprule
\multirow{2}{*}{Accuracy} & {ERM} & {CVaR DRO} & {LfF} & {JTT} & {VICReg} & {SimSiam} & {SimCLR} & {\textbf{DeFund}} \\
{} & {\cmark \xmark} & {\cmark \xmark} & {\cmark \xmark} & {\cmark \xmark} & {\xmark \xmark} & {\xmark \xmark} & {\xmark \xmark} & {\xmark \xmark} \\ 
\midrule
{Worst-group} & {47.2} & {64.4} & {77.2} & {\textbf{81.1}} & {10.2} & {1.1} & {17.1} & {\textbf{77.9}} \\
\midrule
{Average} & {\textbf{95.6}} & {82.5} & {85.1} & {88.0} & {89.0} & {89.0} & {88.9} & {\textbf{89.0}} \\
\bottomrule
\end{tabular}
\end{table}

\begin{table}[htbp]
\caption{(Semi-supervised learning) Worst-group and average accuracy evaluated on CelebA (blonde). Label fraction is set to $10\%$. Each first and second \cmark marker represents whether the model requires information on target class or dataset bias in pretraining stage, respectively.} 
\label{supple table: blonde semi supervised}
\centering
\begin{tabular}{c c c c c c c c}
\toprule
\multirow{2}{*}{Accuracy} & {LNL} & {EnD} & {JTT} & {CVaR DRO} & {ERM} & {SimCLR} & {\textbf{DeFund}} \\
{} & {\cmark \cmark} & {\cmark \cmark} & {\cmark \xmark} & {\cmark \xmark} & {\cmark \xmark} & {\xmark \xmark} & {\xmark \xmark} \\ 
\midrule
{Worst-group ($\%$)} & {40.3} & {41.5} & {79.2} & {49.1} & {30.8} & {12.8} & {\textbf{80.8}} \\
\midrule
{Average ($\%$)} & {\textbf{91.1}} & {91.0} & {91.0} & {91.0} & {89.1} & {89.1} & {90.0} \\
\bottomrule
\end{tabular}
\end{table}

\begin{table}[htbp]
\centering
    \begin{subtable}[h]{0.48\textwidth}
        \centering
        \begin{tabular}{c c c}
            \toprule
            {Methods} & {Worst-group ($\%$)} & {Average ($\%$)} \\
            \midrule
            {JTT} & {70.6} & {86.6} \\
            \midrule
            {\textbf{DeFund}} & {\textbf{75.1}} & {\textbf{94.8}} \\
            \bottomrule
       \end{tabular}
       \caption{Accuracy}
       \label{supp table: subsampled CelebA acc}
    \end{subtable}
    \begin{subtable}[h]{0.48\textwidth}
        \centering
        \begin{tabular}{c c c c}
            \toprule
            \multicolumn{2}{c}{\multirow{2}{*}{}} & \multicolumn{2}{c}{\texttt{Male}} \\
            \cmidrule{3-4}
            \multicolumn{2}{c}{} & 0 & 1 \\
            \midrule
            \multirow{2}{*}{\texttt{Blonde}} & {0} & {1558} & {53483} \\
                                            {} & {1} & {18417} & {1102} \\
            \bottomrule
       \end{tabular}
       \caption{Subsampled CelebA (blonde)} 
       \label{supp table: subsampled CelebA (blonde)}
    \end{subtable}
    \caption{(Semi-supervised learning) (\textbf{a}) Worst-group and average accuracy evaluated on subsampled CelebA (blonde). Label fraction is set to 10$\%$. (\textbf{b}) Number of training samples for each group in subsampled CelebA (blonde). (Original dataset in Table \ref{supple: num celebA (blonde)})}
\label{supple table: subsampled celeba}
\end{table}

\begin{table}[htbp]
\centering
    \begin{subtable}[h]{0.45\textwidth}
        \centering
        \begin{tabular}{c c c}
        \toprule
        {} & Conflict & Unbiased \\
        \midrule
        \rowcolor{Gray}
        SimSiam & 28.15 & 62.63 \\
        \midrule
        + Rank reg & 23.40 & 59.65 \\
        \midrule
        + Upweight & 56.12 & 65.44 \\
        \midrule
        \rowcolor{Gray}
        $\textbf{DeFund}_{\text{Siam}}$ & \textbf{60.37} & \textbf{67.78} \\
        \bottomrule
       \end{tabular}
       \caption{SimSiam}
       \label{supple table: simsiam}
    \end{subtable}
    \begin{subtable}[h]{0.45\textwidth}
        \centering
        \begin{tabular}{c c c}
        \toprule
        {} & Conflict & Unbiased \\
        \midrule
        \rowcolor{Gray}
        VICReg & 32.33 & 64.58 \\
        \midrule
        + Rank reg & 29.73 & 62.08 \\
        \midrule
        + Upweight & 51.19 & 63.41 \\
        \midrule
        \rowcolor{Gray}
        $\textbf{DeFund}_{\text{VIC}}$ & \textbf{53.93} & \textbf{66.31} \\
        \bottomrule
       \end{tabular}
       \caption{VICReg}
       \label{supple table: vicreg}
    \end{subtable}
    
    \caption{Bias-conflict accuracy and unbiased accuracy evaluated on UTKFace (age). Last row corresponds to the full version of proposed framework which upweights misclassified samples identified by biased model. Results are averaged on 4 different random seeds. Accuracy is reported in ($\%$).}
\label{supple table: non-contrastive}
\end{table}

\subsection{Non-contrastive methods}\label{sec: non_CL} 
We provide the results of proposed framework implemented based on non-contrastive methods. Specifically, we leverage SimSiam \citep{chen2021exploring} and VICReg \citep{bardes2021vicreg} as baselines. Table \ref{supple table: non-contrastive} shows that the generalization performance of both baselines can be improved with the proposed debiasing framework. 
% 써야 하는 상황이 있지만, 그럼에도 불구하고 쉽진 않음.

\subsection{Hyperparameter analysis} 
\label{sec: hyperparameter analysis}
While rank regularization biases the representations effectively, we do not argue that it is the optimal form of semantic bottleneck but rather that it highlights the unrecognized importance of controlling effective rank in encoding biased representations. In this regard, we examine the impacts of existing optimization hyperparameters on the effective rank and degree of bias in latent representations. Specifically, we investigated four candidates of bias controller through the lens of effective rank and generalizations: hardness concentration parameter $\beta$ of hard negative sampling \citep{robinson2020contrastive}, temperature $\tau$ in InfoNCE \citep{oord2018representation} loss, strength of $\ell_2$ regularization $\lambda_{\ell_2}$ and the number of training epochs $T$. 

\textbf{Hardness concentration parameter.} Recent works \citep{robinson2020contrastive, cai2020all, tabassum2022hard} stress out the importance of negative examples that are difficult to distinguish from an anchor point. Several recent works propose algorithms on selecting informative negative samples, often controlled by hardness concentration parameter $\beta$ \citep{robinson2020contrastive} coupled with importance sampling. Robinson et al. \cite{robinson2021can} conducted a synthetic simulation showing that increasing $\beta$ makes instance discrimination tasks more difficult, thereby enforcing the encoder to represent more complex features. Thus we aim to examine whether $\beta$ can contribute to learn a debiased representations with real-world dataset. 

\textbf{Temperature.} A recent work on contrastive loss \citep{wang2021understanding} have revealed that temperature $\tau$ can also control the strength of penalties on hard negative samples. Contrastive loss with high temperature turns out to be less sensitive to the hard negative samples \citep{robinson2020contrastive, robinson2021can}, thereby encouraging representations to be locally clustered while the uniformity of features on the hypersphere decreases \citep{wang2020understanding}.
That being said, we hypothesized that the temperature $\tau$ may indirectly affect the effective dimensionality of representations, where large $\tau$ may decrease the effective rank.

\textbf{$\ell_2$ regularization and early-stopping.} Recent studies \cite{sagawa2019distributionally, sagawa2020investigation} underline the importance of regularization for worst-case generalization where the naive upweighting strategy may fail if it is not coupled with strong regularization that prevents deep networks from memorizing upweighted bias-conflicting samples. In this regard, \cite{liu2021just} leverages capacity control techniques, e.g., strong $\ell_2$ regularization or early-stopping, to train complexity-constrained bias-capturing models. We investigate whether such regularizations can serve as a bias controller in self-supervised learning as well.

\begin{figure}[htbp]
\centering
\begin{subfigure}[c]{0.32\textwidth}
\includegraphics[width=\textwidth]{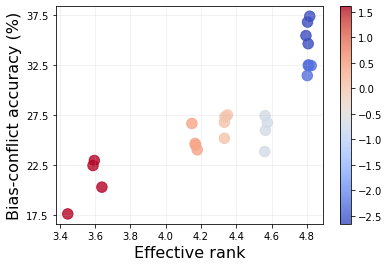} 
\subcaption[c]{$\tau$ on UTKFace (age)}
\label{supple fig: tau (age)}
\end{subfigure}
\begin{subfigure}[c]{0.32\textwidth}
\includegraphics[width=\textwidth]{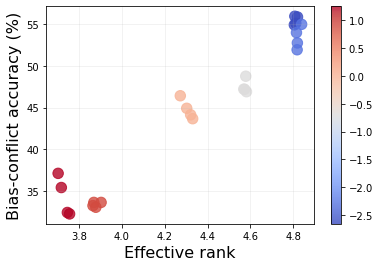} 
\subcaption[c]{$\tau$ on UTKFace (gender)}
\label{supple fig: tau (gender)}
\end{subfigure}
\begin{subfigure}[c]{0.32\textwidth}
\includegraphics[width=\textwidth]{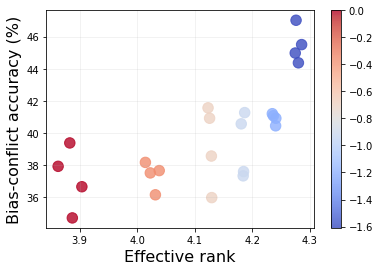} 
\subcaption[c]{$\tau$ on CelebA (makeup)}
\label{supple fig: tau (celeba)}
\end{subfigure}
\begin{subfigure}[c]{0.32\textwidth}
\includegraphics[width=\textwidth]{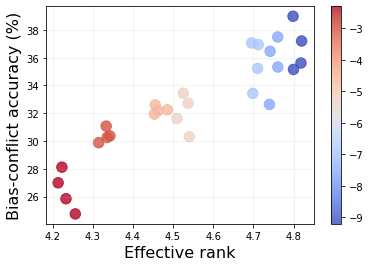} 
\subcaption[c]{$\lambda_{\ell_2}$ on UTKFace (age)}
\label{supple fig: wd (age)}
\end{subfigure}
\begin{subfigure}[c]{0.32\textwidth}
\includegraphics[width=\textwidth]{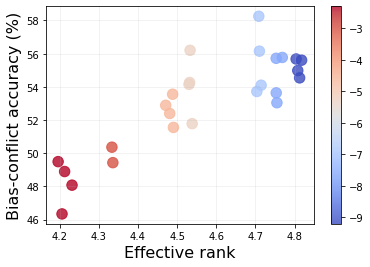} 
\subcaption[c]{$\lambda_{\ell_2}$ on UTKFace (gender)}
\label{supple fig: wd (gender)}
\end{subfigure}
\caption{Analysis on temperature $\tau$ and strength of $\ell_2$ regularization $\lambda_{\ell_2}$. Effective rank and bias-conflict accuracy are measured with varying $\tau$ for (\textbf{a}, \textbf{b}, \textbf{c}), and $\lambda_{\ell_2}$ for (\textbf{d}, \textbf{e}). Standard deviation of bias-aligned accuracy on each experiment is 1.0$\%$, 2.8$\%$, 0.3$\%$, 1.3$\%$ and 1.7$\%$ in order. Performance become quickly degenerated as $\lambda_{\ell_2}$ increases over $0.005$ in CelebA (makeup).}
\end{figure}

\begin{table}[htbp]
\centering
    \begin{subtable}[h]{0.63\textwidth}
        \centering
        \begin{tabular}{c c c c c c}
        \toprule
        {Accuracy} & {$0.01$} & {$0.05$} & {$0.1$} & {$0.15$} & {$1$} \\
        \midrule
        {Conflict} & {35.8} & {36.3} & {37.5} & {37.6} & {36.6} \\
        \midrule
        {Unbiased} & {65.6} & {65.6} & {66.6} & {66.5} & {66.0} \\
        \bottomrule
       \end{tabular}
       \caption{Biased linear evaluation}
       \label{supple table: beta biased eval}
    \end{subtable}
    \begin{subtable}[h]{0.35\textwidth}
        \centering
        \begin{tabular}{c c c}
        \toprule
        {} & SimCLR & $\beta$=0.1 \\
        \midrule
        Conflict & 62.0 & 64.2 \\
        \midrule
        Unbiased & 78.9 & 80.7 \\
        \bottomrule
       \end{tabular}
       \caption{Debiased linear evaluation}
       \label{supple table: beta debiased eval}
    \end{subtable}
    \caption{Results of controlling concentration parameter $\beta$ on UTKFace (age). Accuracy is reported in ($\%$). (\textbf{a}): Accuracy of linear evaluation without upweighting bias-conflicting samples. Each value in top row indicates $\beta$ used in pretraining. (\textbf{b}) Accuracy of linear evaluation with upweighting ground-truth bias-conflicting samples. Both models use $\lambda_{up}=10$.}
\label{supple table: beta}
\end{table}

\textbf{Results.} 
We evaluate each knob on generalizations with SimCLR. Table \ref{supple table: earlystop} and \ref{supple table: beta biased eval} show that impacts of both early-stopping and concentration parameter $\beta$ on generalizations are marginal, in contrast to the observations reported in supervised learning or synthetic simulations \citep{robinson2021can}. However, it still remains unclear whether the debiased representations can be encoded by controlling $\beta$. It is because the model may reach a biased solution even though it encodes debiased representations, if most samples in linear evaluation are bias-aligned, as discussed in the main paper. To preclude such confounding relationships, we conduct debiased linear evaluation with upweighting ground-truth bias-conflicting samples. Table \ref{supple table: beta biased eval} and \ref{supple table: beta debiased eval} show that there was no significant difference in the performance gain of $\beta$ in biased and debiased linear evaluation, which implies that $\beta$ is not enough to fully debias representations.

Despite the failure of learning debiased representations with controlling $\beta$, biased representations can be learned by controlling temperature $\tau$, and strength of $\ell_2$ regularization in some cases. Figure \ref{supple fig: tau (age)}, \ref{supple fig: tau (gender)} and \ref{supple fig: tau (celeba)} show that effective rank, temperature and bias-conflicting accuracy are highly correlated each other in both UTKFace and CelebA. It implies that the effective rank can serve as a metric of generalization performance and degree of bias in representations. While temperature control cannot be generalized to several non-contrastive learning methods \citep{chen2021exploring, bardes2021vicreg, zbontar2021barlow}, this results imply that the temperature may serve as an effective bias controller for contrastive learning methods using InfoNCE loss. Moreover, stronger-than-typical $\ell_2$ regularization also limits the effective rank and bias-conflict accuracy to some extent in UTKFace (Figure \ref{supple fig: wd (age)} and \ref{supple fig: wd (gender)}), while it fails to do so in CelebA.

This series of observations afford us a novel insight that both explicit (rank regularization) and implicit (temperature control, strong $\ell_2$ regularization) methods offer a way to train biased representations. However, it still remains unclear how to directly learn \textit{debiased} representations. While increasing temperature or reducing effective rank bias representations, inverse does not always hold; Abnormally small temperatures cause the contrastive loss only focus on the nearest one or two samples, which heavily degenerates the performance \citep{wang2021understanding}. Moreover, we found that explicit decorrelation of feature components in SimCLR does not lead to debiased representations (not shown in figure). %The failure of Implicit Feature Modification \citep{robinson2021can} and hard negative sampling in real-world debiasing is potentially attributed to  

To sum up, we provide useful recipes on learning biased representations, where rank regularization is mainly discussed in the main paper due to its intuitive insights, good performance and broad applicability. We hope these discussions facilitate in-depth studies about advanced algorithms on learning both biased and debiased representations in unsupervised manner.

\begin{table}[htbp]
\caption{Results of early-stopping on UTKFace. We denote $T$ as the number of training epochs. } 
\label{supple table: earlystop}
\centering
\begin{tabular}{c c c c c c c}
\toprule
{Attribute} & {Accuracy} & {$T=5$} & {$T=10$} & {$T=15$} & {$T=20$} & {$T=25$} \\
\midrule
\multirow{2}{*}{Age}  & 
{Bias-conflict ($\%$)} & {31.6} & {33.0} & {32.4} & {32.8} & {32.8} \\
\cmidrule{2-7}
& {Unbiased ($\%$)} & {63.3} & {64.1} & {63.6} & {63.7} & {63.7} \\
\midrule
\multirow{2}{*}{Gender} & 
{Bias-conflict ($\%$)} & {54.6} & {54.0} & {53.5} & {53.4} & {54.5} \\
\cmidrule{2-7}
& {Unbiased ($\%$)} & {72.1} & {72.0} & {71.8} & {72.2} & {72.7} \\
\bottomrule
\end{tabular}
\end{table}

\section{Experimental setup}
\label{sec: experimental setup}
\subsection{Datasets} 
\begin{figure}[htbp]
\centering
\includegraphics[width=\textwidth]{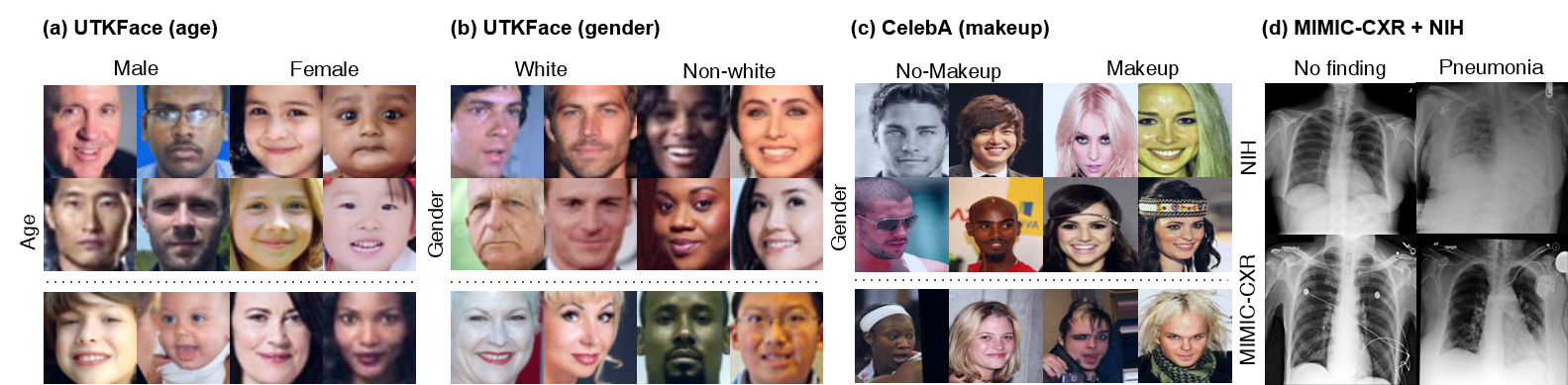} 
\caption{Example images of datasets. Top-row annotations refer to the target attributes, i.e. \texttt{Gender, Race, HeavyMakeup} and \texttt{Pneumonia}, while the left-side annotations refer to the bias attributes, i.e. \texttt{Age, Gender} and data source, respectively. For (\textbf{a}), (\textbf{b}), and (\textbf{c}), the images above the dotted line denote the bias-aligned samples, while the ones below the dotted line are the bias-conflicting samples.}\label{supple fig: examples}
\end{figure}

We mainly evaluate our debiasing framework on MultiCMNIST \citep{li2022discover}, MIMIC-CXR + NIH \citep{li2023partition}, UTKFace \citep{zhang2017age} and CelebA \citep{liu2015deep} in which several prior works has observed poor generalization performance due to spurious correlations. Example images are presented in Figure \ref{supple fig: examples}.

\textbf{MultiCMNIST.} It is worth noting that existing off-the-shelf synthetic biased datasets often fail to account for real-world scenarios in which multiple bias attributes can coexist simultaneously. To address this limitation, the work by \cite{li2022discover} introduces the innovative Multi-Color MNIST (MultiCMNIST) dataset, designed to emulate complex real-world multi-bias scenarios. Specifically, there are two bias attributes, namely \texttt{left color} and \texttt{right color}, where we set bias ratio=99\% for the left color and bias ratio=95\% for the right color.

\textbf{MIMIC-CXR + NIH.} The dataset discussed here serves as a poignant example of spurious correlations within medical imaging datasets. In such datasets, machine learning classifiers may struggle to discern the true underlying pathological indicators, such as the presence of pneumonia, often relying on spurious radiographic features tied to variations in data acquisition procedures \citep{degrave2021ai}. To simulate spurious correlations in medical imaging dataset, we mix MIMIC-CXR \citep{johnson2019mimic} and NIH \citep{wang2017chestx} datasets into a MIMIC-CXR + NIH dataset following \cite{li2023partition}. The original NIH contains 50500 \texttt{no finding} and 876 \texttt{pneumonia} training images, while the original MIMIC-CXR has 10145 \texttt{no finding} and 7209 \texttt{pneumonia}
training images. Given the scarcity of \texttt{pneumonia} images in the NIH dataset, we curate the MIMIC-CXR + NIH dataset by primarily extracting \texttt{pneumonia} images from MIMIC-CXR and \texttt{no finding} images from NIH. In MIMIC-CXR + NIH, the target categories are \texttt{no finding} and \texttt{pneumonia}, and the biases come from two data sources. It contains 8000 training images with a bias ratio of 0.9, 250 unbiased validation images, and 250 unbiased test images. 

\textbf{UTKFace.} We first consider UTKFace dataset which is consist of human face images with varying \texttt{Race, Gender} and \texttt{Age} attributes. For each sensitive attribute, we categorize all samples into two groups. Specifically, for label associated with age, we assign 1 to samples with $\age \leq 10$, and 0 to samples with $\age \geq 20$ following \citep{hong2021unbiased}. For label associated with race, we assign 1 to samples with $\race \neq \white$, e.g., Black, Indian and Asian, and 0 to samples with $\race = \white$. For label associated with gender, we assign 1 to female, and 0 to male. Based on this settings, we conduct binary classifications using (\texttt{Gender, Age}) and (\texttt{Race, Gender}) as (target, spurious) attribute pairs. Following \cite{hong2021unbiased}, we construct a biased dataset by randomly truncating a portion of samples, where roughly 90$\%$ of samples are bias-aligned in our setting. Pixel resolutions and batch size are $64 \times 64$ and 256, respectively.

\textbf{CelebA.} For CelebA, we consider (\texttt{HeavyMakeup}, \texttt{Male}) and (\texttt{Blonde Hair}, \texttt{Male}) as (target, spurious) attribute pairs, following \cite{nam2020learning, hong2021unbiased, sagawa2019distributionally}. Pixel resolutions and batch size are $256 \times 256$ and 128, respectively. The exact number of samples for each prediction task is summarized in Table \ref{supple table: num data}. 

\begin{table}[htbp]
\centering
    \begin{subtable}[h]{0.19\textwidth}
        \centering
        \resizebox{0.99\textwidth}{!}{
        \begin{tabular}{c c c c}
            \toprule
            \multicolumn{2}{c}{\multirow{2}{*}{}} & \multicolumn{2}{c}{\texttt{A}} \\
            \cmidrule{3-4}
            \multicolumn{2}{c}{} & 0 & 1 \\
            \midrule
            \multirow{2}{*}{\texttt{G}} & {0} & {8229} & {822} \\
                                            {} & {1} & {134} & {1346} \\
            \bottomrule
       \end{tabular}
       }
       \caption{UTKFace (\texttt{A})} 
    \end{subtable}
    \begin{subtable}[h]{0.19\textwidth}
        \centering
        \resizebox{0.99\textwidth}{!}{
        \begin{tabular}{c c c c}
            \toprule
            \multicolumn{2}{c}{\multirow{2}{*}{}} & \multicolumn{2}{c}{\texttt{G}} \\
            \cmidrule{3-4}
            \multicolumn{2}{c}{} & 0 & 1 \\
            \midrule
            \multirow{2}{*}{\texttt{R}} & {0} & {4354} & {534} \\
                                            {} & {1} & {435} & {5344} \\
            \bottomrule
       \end{tabular}
       }
       \caption{UTKFace (\texttt{G})} 
    \end{subtable}
    \begin{subtable}[h]{0.19\textwidth}
        \centering
        \resizebox{0.99\textwidth}{!}{
        \begin{tabular}{c c c c}
            \toprule
            \multicolumn{2}{c}{\multirow{2}{*}{}} & \multicolumn{2}{c}{\texttt{M}} \\
            \cmidrule{3-4}
            \multicolumn{2}{c}{} & 0 & 1 \\
            \midrule
            \multirow{2}{*}{\texttt{H}} & {0} & {25789} & {54460} \\
                                            {} & {1} & {49804} & {163} \\
            \bottomrule
       \end{tabular}
       }
       \caption{CelebA (\texttt{H})}
       \label{supple: num celebA (makeup)}
    \end{subtable}
    \begin{subtable}[h]{0.19\textwidth}
        \centering
        \resizebox{0.99\textwidth}{!}{
        \begin{tabular}{c c c c}
            \toprule
            \multicolumn{2}{c}{\multirow{2}{*}{}} & \multicolumn{2}{c}{\texttt{M}} \\
            \cmidrule{3-4}
            \multicolumn{2}{c}{} & 0 & 1 \\
            \midrule
            \multirow{2}{*}{\texttt{B}} & {0} & {57214} & {53483} \\
                                            {} & {1} & {18417} & {1102} \\
            \bottomrule
       \end{tabular}
       }
       \caption{CelebA (\texttt{B})}
       \label{supple: num celebA (blonde)}
    \end{subtable}
    \begin{subtable}[h]{0.19\textwidth}
        \centering
        \resizebox{0.99\textwidth}{!}{
        \begin{tabular}{c c c c}
            \toprule
            \multicolumn{2}{c}{\multirow{2}{*}{}} & \multicolumn{2}{c}{\texttt{Data}} \\
            \cmidrule{3-4}
            \multicolumn{2}{c}{} & NIH & MIMIC \\
            \midrule
            \multirow{2}{*}{\texttt{P}} & {0} & {3600} & {400} \\
                                            {} & {1} & {400} & {3600} \\
            \bottomrule
       \end{tabular}
       }
       \caption{MIMIC+NIH}
       \label{supple: num mimic}
    \end{subtable}
    \caption{Number of training samples for each prediction task. \texttt{A} for \texttt{Age}, \texttt{G} for \texttt{Gender}, \texttt{R} for \texttt{Race}, \texttt{M} for \texttt{Male}, \texttt{H} for \texttt{HeavyMakeup}, \texttt{B} for \texttt{Blonde Hair}, and \texttt{P} for \texttt{Pneumonia}.}
\label{supple table: num data}
\end{table}

\subsection{Rank reduction \& regularization analysis} 

\textbf{CMNIST \& MultiCMNIST.} For CMNIST, we use a simple convolutional network with three convolution layers as a counterpart of ResNet-18, with feature map dimensions of 64, 128, and 256, each followed by a ReLU activation and a batch normalization. The convolutional network is trained for 2000 iterations using SGD optimizer with initial learning rate 0.1 and decaying by 0.1 for every 600 iterations, following \cite{zhang2021can}. For a MultiCMNIST, the experimental settings including neural architecture and optimizer follow the original paper \citep{li2022discover} for a fair comparison.

\textbf{CIFAR10-C and Waterbirds.} For CIFAR10-C and Waterbirds in Figure \ref{2 fig: waterbirds group}, we use ResNet-18 and ResNet-50 with pretrained weights provided in PyTorch torchvision implementations, respectively. ResNet-18 is trained for 10000 iterations using the Adam optimizer with learning rate 0.001. After training, misclassified training samples are identified as the bias-conflicting samples as in Table \ref{2 table: precision}. Following the official implementation of JTT, ResNet-50 is trained for 300 epochs, and early-stopped with referring to the validation accuracy, using SGD optimizer with learning rate 0.0001.

\textbf{Hyperparameters.} In Table \ref{2 table: precision}, $\lambda_{reg}=35$ and $\lambda_{reg}=20$ are used for CMNIST and CIFAR-10C, respectively. In Table \ref{2 table: precision_jtt}, $\lambda_{reg}=10$ is used.

\subsection{Debiasing experiments}

\textbf{Architecture details.} We use ResNet-18 back-bone architecture with pretrained weights provided in in PyTorch \texttt{torchvision} implementations. For projection networks in SimCLR, we use the MLP consists of one hidden layer with feature dimension of 512, followed by a ReLU activation.
We employ a single linear classifier in downstream tasks for all self-supervised learning methods.

\textbf{Training details.} For MIMIC-CXR+NIH, both biased and main classifiers are trained by using Adam optimizer with learning rate of 0.0003. Biased and main classifiers are trained for 5 and 100 epochs, respectively. For a rank regularization, $\lambda_{reg}=10$ is used. For a upweighting, $\lambda_{up}=5$ is used with $\lambda_{\ell_2}=0.0005$. 

Both biased and main encoders are pretrained for 100 epochs on UTKFace, and 20 epochs on CelebA, by using Adam optimizer with learning rate of 0.0003. Cosine annealing scheduling \citep{loshchilov2016sgdr} is leveraged with warmup for the first 20 epochs on UTKFace, and 4 epochs for CelebA. 

For biased encoders, we apply rank regularization with using $\lambda_{reg}$ of 0.3, 0.5, 0.01 and 0.03 for UTKFace (age), UTKFace (gender), CelebA (makeup) and CelebA (blonde), respectively. This values are selected by tuning $\lambda_{reg} \in \{0.0, 0.1, 0.3, 0.5, 1.0\}$ for UTKFace and $\lambda_{reg} \in \{0.0, 0.01, 0.02, 0.03, 0.05\}$ for CelebA. Specifically, we report the results of model with highest worst-group accuracy (for CelebA (blonde)), or bias-conflicting test accuracy over those with improved unbiased test accuracy compared to the SimCLR baseline. Same values are consistently used for upweighting in ablation study (Table \ref{4. table: ablation}). To emphasize the contribution of rank regularization, we do not control any other parameters, e.g., strength of $\ell_2$ regularization, temperature $\tau$, or number of training epochs. Specifically, we fix $\tau=0.07$ and $\lambda_{\ell_2}=0.0001$ for every experiment.

After pretraining, we conduct either linear evaluation or finetuning with using $\lambda_{up}$ of 10, 5, 8 and 15 for UTKFace (age), UTKFace (gender), CelebA (makeup) and CelebA (blonde), respectively. For UTKFace and CelebA (makeup), these values are selected by tuning $\lambda_{up} \in \{5, 8, 10\}$ using the above-mentioned decision rules, where $\lambda_{up} \in \{5, 8, 10, 15\}$ is compared for CelebA (blonde). Same values are consistently used in ablation study (Table \ref{4. table: ablation}).  For linear evaluation, we train a linear classifier on top of pretrained main encoder for 3000 iterations on UTKFace, and 5000 iterations on CelebA, with using learning rate of 0.0003 and upweighting identified bias-conflicting samples. For semi-supervised learning, we finetune the whole main model for 5000 iterations, with using SGD optimizer, momentum of 0.9, $\lambda_{\ell_2}=0.1$, learning rate of 0.0001, and $\lambda_{up}=8, 15$ for CelebA (makeup) and CelebA (blonde), respectively. 

\textbf{Data augmentations.} Following SimCLR, we generate multiviewed batch with random augmentations of (a) random resized crop with setting the scale from 0.2 to 1, (b) random horizontal flip with the probability of 0.5, (c) random color jitter (change in brightness, contrast, and saturation) with the probability of 0.8 and scale of 0.4, (d) random gray scaling with the probability of 0.2. In linear evaluation and finetuning, we only apply random horizontal flip. Same augmentation pipeline is applied to both SimSiam and VICReg.

\textbf{Baselines.} For a fair comparison, we tune hyperparameters of other baselines using the same ResNet-18 back-bone architecture. We use the official implementation of JTT which also includes that of CVaR DRO. Other baselines are reproduced by ourselves with referring to original papers. LNL is trained for 20 epochs on UTKFace, and 40 epochs on CelebA and MIMIC-CXR + NIH, with using Adam optimizer and learning rate of 0.001. For EnD, we set the multipliers $\alpha$ for disentangling and $\beta$ for entangling to 1. For JTT, we tune the upweighting factor $\lambda_{up} \in \{20, 50, 80\}$ and number of training epochs $T \in \{30, 40, 50\}$, following the original paper. For CVaR DRO, we tune the size of the worst-case
subpopulation $\alpha \in \{0.1, 0.2, 0.5\}$. For SimSiam and VICReg, the architectures for the additional layers followed the official implemenation of each method, where the hyperparameters for the training is identical to the SimCLR case. For \ref{sec: non_CL}, $\lambda_{reg}=0.001$ for $\textbf{DeFund}_{\text{Siam}}$ and $\lambda_{reg}=0.1$ for $\textbf{DeFund}_{\text{VIC}}$.

\end{document}